\documentclass[sigconf]{acmart}
\usepackage{graphicx}
\usepackage{caption}
\usepackage{subcaption}
\usepackage{algorithmic}
\usepackage{algorithm}
\usepackage{graphicx}
\usepackage{textcomp}
\usepackage{xcolor}
\usepackage{bbm}
\usepackage{algorithm}
\usepackage{caption}
\usepackage{subcaption}
\usepackage{enumitem}
\usepackage{booktabs}
\usepackage{amsfonts}
\usepackage{amsmath}
\usepackage{hyperref}
\usepackage{multirow}
\usepackage{enumitem}
\usepackage{makecell}
\newcommand{\indep}{\perp \!\!\! \perp}
\usepackage{hhline}

\newtheorem{definition}{Definition}[]
\newtheorem{problem}{Problem}[]

\newcount\DraftStatus  
\usepackage{color}

\ifnum\DraftStatus=1
\usepackage[draft,colorinlistoftodos,color=orange!30]{todonotes}
\else
\usepackage[disable,colorinlistoftodos,color=blue!30]{todonotes}
\fi

\AtBeginDocument{%
  \providecommand\BibTeX{{%
    \normalfont B\kern-0.5em{\scshape i\kern-0.25em b}\kern-0.8em\TeX}}}

\setcopyright{acmcopyright}
\copyrightyear{2018}
\acmYear{2018}
\acmDOI{10.1145/1122445.1122456}

\acmConference[Woodstock '18]{Woodstock '18: ACM Symposium on Neural
  Gaze Detection}{June 03--05, 2018}{Woodstock, NY}
\acmBooktitle{Woodstock '18: ACM Symposium on Neural Gaze Detection,
  June 03--05, 2018, Woodstock, NY}
\acmPrice{15.00}
\acmISBN{978-1-4503-XXXX-X/18/06}

\begin{document}

\title{Be Causal: De-biasing Social Network Confounding in Recommendation}

\author{Qian Li}
\authornotemark[1]
\authornote{Both authors contributed equally to this research.}
\email{qian.li@uts.edu.au}
\affiliation{%
  \institution{University of Technology, Sydney}
  \streetaddress{P.O. Box 123}
  \city{Broadway}
  \state{NSW 2007}
  \country{Australia}
}

\author{Xiangmeng Wang}
\authornotemark[1]
\email{xiangmeng.wang@student.uts.edu.au}
\affiliation{%
  \institution{University of Technology, Sydney}
  \streetaddress{P.O. Box 123}
  \city{Broadway}
  \state{NSW 2007}
  \country{Australia}
}
\author{Guandong Xu}
\authornotemark[1]
\email{guandong.xu@uts.edu.au}
\affiliation{%
  \institution{University of Technology, Sydney}
  \streetaddress{P.O. Box 123}
  \city{Broadway}
  \state{NSW 2007}
  \country{Australia}
}

\begin{abstract}
In recommendation systems, the existence of the missing-not-at-random (MNAR) problem results in the selection bias issue, degrading the recommendation performance ultimately.
A common practice to address MNAR is to treat missing entries from the so-called ``exposure'' perspective, i.e., modeling how an item is exposed (provided) to a user.
Most of the existing approaches use heuristic models or re-weighting strategy on observed ratings to mimic the missing-at-random setting.
However, little research has been done to reveal how the ratings are missing from a causal perspective.
To bridge the gap, we propose an unbiased and robust method called DENC (\emph{De-bias Network Confounding in Recommendation}) inspired by confounder analysis in causal inference.
In general, DENC provides a causal analysis on MNAR from both the inherent factors (e.g., latent user or item factors) and auxiliary network's perspective.
Particularly, the proposed exposure model in DENC can control the social network confounder meanwhile preserves the observed exposure information. We also develop a deconfounding model through the balanced representation learning to retain the primary user and item features, which enables DENC generalize well on the rating prediction.
Extensive experiments on three datasets validate that our proposed model outperforms the state-of-the-art baselines.

\end{abstract}



\maketitle

\section{Introduction}

Recommender systems aim to handle information explosion meanwhile to meet users' personalized interests, which have received extensive attention from both research communities and industries.
The power of a Recommender system highly relies on whether the observed user feedback on items ``correctly'' reflects the users’ preference or not.
However, such feedback data often contains only a small portion of observed feedback (e.g., explicit ratings), leaving a large number of missing ratings to be predicted. 
To handle the partially observed feedback, a common assumption for model building is that the feedback is missing at random (MAR), i.e., the probability of a rating to be missing is independent of the value. When the observed data follows the MAR, using only the observed data via statistical analysis methods can yield ``correct'' prediction without introducing bias~\cite{marlin2009collaborative,lim2015top}.
However, this MAR assumption usually does not hold in reality and the missing pattern exhibits \emph{missing not at random} (MNAR) phenomenon. Generally, MNAR is related to selection bias.
For instance, in movie recommendation, instead of randomly choosing movies to watch, users are prone to those that are highly recommended, while in advertisement recommendation, whether an advertisement is presented to a user is purely subject to the advertiser's provision, rather than at random. 
In these scenarios, the missing pattern of data mainly depends on whether the users are exposed to the items, and consequently, the ratings in fact are \emph{missing not at random} (MNAR)~\cite{he2016fast}. 
These findings shed light on the origination of selection bias from MNAR ~\cite{sportisse2020imputation}. 
Therefore the selection bias cannot be ignored in practice and it has to be modeled properly in order for reliable recommendation prediction. How to model the missing data mechanism and debias the rating performance forms up the main motivation of this research.

\noindent\textbf{Existing MNAR-aware Methods}

There are abundant methods for addressing the MNAR problem on the implicit or explicit feedback.
For implicit feedback, traditional methods~\cite{hu2008collaborative} take the uniformity assumption that assigns a uniform weight to down-weight the missing data, assuming that each missing entry is equally likely to be negative feedback.  
This is a strong assumption and limits models’ flexibility for real applications.
Recently, researchers tackle MNAR data directly through simulating the generation of the missing pattern under different heuristics~\cite{hernandez2014probabilistic}.
Of these works, probabilistic models are presented as a proxy to relate missing feedback to various factors, e.g., item features. 
For explicit feedback, a widely adopted mechanism is to exploit the dependencies between rating missingness and the potential ratings (e.g., 1-5 star ratings)~\cite{koren2015advances}. That is, high ratings are less likely to be missing compared to items with low ratings.
However, these paradigm methods involve heuristic alterations to the data, which are neither empirically verified nor theoretically proven~\cite{saito2020asymmetric}. 

A couple of methods have recently been studied for addressing MNAR~\cite{hernandez2014probabilistic,liang2016causal,schnabel2016recommendations} by treating missing entries from the so-called ``exposure'' perspective, i.e., indicating whether or not an item is exposed (provided) to a user.
For example, ExpoMF resorts modeling the probability of \emph{exposure}~\cite{hernandez2014probabilistic}, and up-weighting the loss of rating prediction with high \emph{exposure} probability.
However, ExpoMF can lead to a poor prediction accuracy for rare items when compared with popular items.
Likewise, recent works~\cite{liang2016causal,schnabel2016recommendations} resort to \emph{propensity score} to model \emph{exposure}.
The \emph{propensity score} introduced in causal inference indicates \emph{the probability that a subject receiving the treatment or action}.
Exposing a user to an item in a recommendation system is analogous to exposing a subject to a treatment. 
Accordingly, they adopt \emph{propensity score} to model the \emph{exposure} probability and re-weight the prediction error for each observed rating with the inverse \emph{propensity score}.
The ultimate goal is to calibrate the MNAR feedbacks into missing-at-random ones that can be used to guide unbiased rating prediction.

Whilst the state-of-the-art propensity-based methods are validated to alleviate the MNAR problem for recommendation somehow,
they still suffer from several major drawbacks:
1) they merely exploit the user/item latent vectors from the ratings for mitigating MNAR, 
but fail to disentangle different causes for MNAR from a causal perspective;
2) technically, they largely rely on propensity score estimation to mitigate MNAR problem; the performance is sensitive to the choice of propensity estimator~\cite{wang2019doubly},
which is notoriously difficult to tune.
 \begin{figure}[!htb]
 \includegraphics[width=0.8\linewidth]{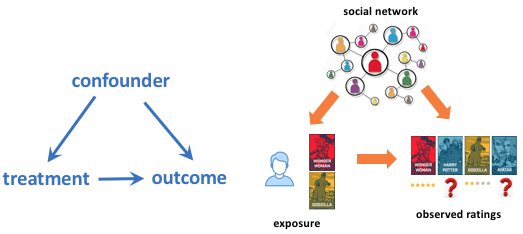}  
\caption{The causal view for MNAR problem: \emph{treatment} and \emph{outcome} are terms in the theory of causal inference, 
which denote an action taken (e.g.,\emph{exposure}) and its result (e.g., \emph{rating}), respectively. The \emph{confounder} (e.g., \emph{social network}) is the common cause of treatment and outcome.}
\label{fig:example}
\end{figure}

\noindent\textbf{The proposed approach}

\begin{figure}[!htb]
 \includegraphics[width=0.8\linewidth]{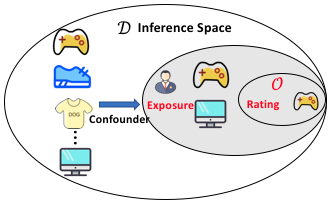}  
\caption{The training space of conventional recommendation models is the observed rating space $\mathcal{O}$, whereas the inference space is the entire exposure space $\mathcal{D}$.
The discrepancy of data distribution between $\mathcal{O}$ and $\mathcal{D}$ leads to selection bias in conventional recommendation models.}
\label{fig:seb}
\end{figure}

To overcome these obstacles, in contrast, we aim to address the fundamental MNAR issue in recommendation from a novel causal inference perspective, to attain a robust and unbiased rating prediction model. 
From a causal perspective, we argue that the selection bias (i.e., MNAR) in the recommendation system is attributed to the presence of \emph{confounders}. As explained in Figure~\ref{fig:example}, \emph{confounders} are factors (or variables) that affect both
the treatment assignments (exposure) and the outcomes (rating). 
For example, friendships (or social network) can influence both users’ choice of movie watching and their further ratings. Users who choose to watch the movie are more likely to rate than those who do not. So, \emph{the social network is indeed a confounding factor that affects which movie the user is exposed to and how the user rates the movie}.
The confounding factor results in a \emph{distribution discrepancy between the partially observed ratings and the complete ratings} as shown in Figure~\ref{fig:seb}. 
Without considering the distribution discrepancy, the rating model trained on the observed ratings fails to generalize well on the unobserved ratings.
With this fact in mind, our idea is to analyze the confounder effect of social networks on rating and exposure, and in turn, fundamentally alleviate the MNAR problem to predict valid ratings.  

In particular, we attempt to study the MNAR problem in recommendation from a causal view and propose an unbiased and robust method called DENC (\emph{De-bias Network Confounding in Recommendation}).
To sufficiently consider the selection bias in MNAR, we model the underlying factors (i.e., inherent user-item information and social network) that can generate observed ratings. 
In light of this, as shown in Figure~\ref{fig:frame}, we construct a causal graph based recommendation framework by disentangling three determinants for the ratings, i.e., 
\emph{inherent factors},
\emph{confounder} and \emph{exposure}. 
Each determinant accordingly corresponds to one of three specific components in DENC: 
\emph{deconfonder model},
\emph{social network confounder} and \emph{exposure model},
all of which jointly determine the rating outcome.

In summary, the key contributions of this research are as follows:

\begin{itemize}
    \item Fundamentally different from previous works, DENC is the first method for the unbiased rating prediction through disentangling determinants of selection bias from a causal view. 
    \item The proposed \emph{exposure model} is capable of revealing the exposure assignment and accounting for the confounder factors derived from the \emph{social network confounder}, which thus remedies selection bias in a principled manner.
    \item We develop a \emph{deconfonder model} via the balanced representation learning that embeds inherent factors independent of the exposure, therefore mitigating the distribution discrepancy between the observed rating and inference space.
    \item We conduct extensive experiments to show that our DENC method outperforms state-of-the-art methods.
    The generalization ability of our DENC is also validated by verifying different degrees of confounders.
\end{itemize}

\section{Related Work\label{related work}}

\subsection{MNAR-aware Methods}
\subsubsection{Traditional Heuristic Models} 
Early works on explicit feedback formulate a recommendation a rating prediction problem in which the large volume of unobserved ratings (i.e., missing data) is assumed to be extraneous for user preference~\cite{hu2008collaborative}. 
Following this unreliable assumption, numerous recommenders have been developed including basic matrix factorization based-recommenders~\cite{rendle2008online} and sophisticated ones such as SVD++~\cite{koren2015advances}.
As statistical analysis with missing data techniques, especially MNAR proposition, find widespread applications, there is much interest in understanding its impacts on the recommendation system. 
Previous research has shown that for explicit-feedback recommenders, users’ ratings are MNAR~\cite{marlin2009collaborative}.
Marlin and Zemel~\cite{marlin2009collaborative} first study the effect of violating MNAR assumption in recommendation methods; they propose statistical models to address MNAR problem of missing ratings based on the heuristic that users are more likely to supply ratings for items that they do like.
Another work of ~\cite{hernandez2014probabilistic} also has focused on addressing MNAR problem; they propose a probabilistic matrix factorization model (ExpoMF) for collaborative filtering that learns from observation data.
However, these heuristic paradigm methods are neither empirically verified nor theoretically proven~\cite{schnabel2016recommendations,liang2016causal}.
\subsubsection{Propensity-based Model}
The basic idea of propensity scoring methods is to turn the outcomes of an observational study into a pseudo-randomized trial by re-weighting samples, similarly to importance sampling.
Typically, using Inverse Propensity Weighting (IPW) estimator, Liang~\cite{liang2016causal} proposes a framework consisted of one exposure model and one preference model. 
The exposure model is estimated by a Poisson factorization,
then preference model is fit with weighted click data, where each click is weighted by the inverse of exposure and be used to alleviate popularity bias.
Based on Self Normalized Inverse Propensity Scoring (SNIPS) estimator,
the model in \cite{schnabel2016recommendations} are developed either directly through observed ratings of a missing-completely-at-random sample estimated by SNIPS or indirectly through user and item covariates.
These works re-weight the observational click data as though it came from an ``experiment'' where users are randomly shown items. 
Thus, the measurement is still adopting re-weighting strategies to mimic the missing-completely-at-random like most of the heuristic models do~\cite{yang2018unbiased}.
Besides, these works are sensitive to the choice of propensity score estimators~\cite{wang2019doubly}.
In contrast, our work relies solely on the observed ratings: we do not require ratings from a gold-standard randomized exposure estimation and nor do we use external covariates; moreover, we consider another important bias in the recommendation scenario, namely, social counfounding bias. 
\subsection{Social Network-based Methods}
The effectiveness of social networks has been proved by a vast amount of social recommenders. 
Purushotham~\cite{purushotham2012collaborative} has explored how traditional factorization methods can exploit network connections; this brings the latent preferences of connected users closer to each other, reflecting that friends have similar tastes. 
Other research has included social information directly into various collaborative filtering methods. TrustMF~\cite{yang2016social} adopts collaborative filtering to map users into low-dimensional latent feature spaces in terms of their trust relationship; the remarkable performance of the proposed model reflects individuals among a social network will affect each other while reviewing items. 
SocialMF~\cite{jamali2010matrix} incorporates trust propagation into the matrix factorization model, which assumes the factors of every user are dependent on the factor vectors of his/her direct neighbors in the social network. However, despite the remarkable contribution of social network information in various recommendation methods, it has not been utilized in controlling for confounding bias of causal inference-based recommenders yet.
\section{DENC Method}

\subsection{Notations}
We first give some preliminaries of our method and used notation. 
Suppose we have $m \times n$ rating matrix $Y\in\mathbb{R}^{m \times n} =[\dot{y}_{ui}]$ describing the numerical rating of $m$ users on $n$ items. Let \(U = \{u_{1},u_{2},...,u_{n}\}\) and \(I = \{i_{1},i_{2},...,i_{m}\}\) be the set of users and items respectively. For each user-item pair, we use \(a_{ui}\) to indicate whether user \(u\) has been exposed to item \(i\) and $a_{ui}\in \{0,1\}$.
We use \(y_{ui}\) to represent the rating given by \(u\) to item \(i\). 

\subsection{A Causal Inference Perspective}
Viewing recommendation from a causal inference perspective, we argue that exposing a user to an item in recommendation is an intervention analogous to exposing a patient to a treatment in a medical study. Following the potential outcome framework in causal inference~\cite{rubin1974estimating}, we reformulate the rating prediction as follows.
\begin{problem}[Causal View for Recommendation]
For every user-item pair with a binary exposure $a_{ui}\in\{0,1\}$, there are two potential rating outcomes $y_{ui}(0)$ and $y_{ui}(1)$. We aim to \emph{estimating the ratings had all movies been seen by all users}, i.e., estimate $y_{ui}(1)$ (i.e., $y_{ui}$) for all $u$ and $i$. 
\label{th:pof}
\end{problem}

\begin{figure}[!htb]
 \includegraphics[width=0.75\linewidth]{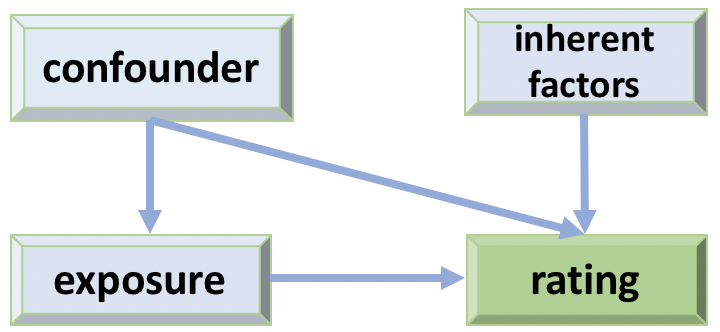}  
\caption{The causal graph for recommendation.}
\label{fig:underlying}
\end{figure}
As we can only observe the outcome $y_{ui}(1)$ when the user $u$ is exposed by the item $i$, i.e., $a_{ui}=1$,  
we target at the problem that \emph{what would happen to the unobserved rating $y_{ui}$ if we set exposure variable by setting $a_{ui}=1$}. 
In our settings, the confounder derived from the social network among users are denoted as a common cause that affects the exposure $a_{ui}$ and outcome $y_{ui}$. We aim to disentangle the underlying factors in observation ratings and social networks as shown in Figure~\ref{fig:underlying}. 
The intuition behind Figure~\ref{fig:underlying} is that the observed rating outcomes are generated as a result of both inherent and confounding factors. The inherent factors refer to the user preferences and inherent item properties, and auxiliary factors are the confounding factors from the social network.
By disentangling determinants that cause the observed ratings, we can account for effects separately from the selection bias of confounders and the exposure, which ensures to attain an unbiased rating estimator with superior generalization ability.

\begin{figure}[!htb]
 \includegraphics[width=1.0\linewidth]{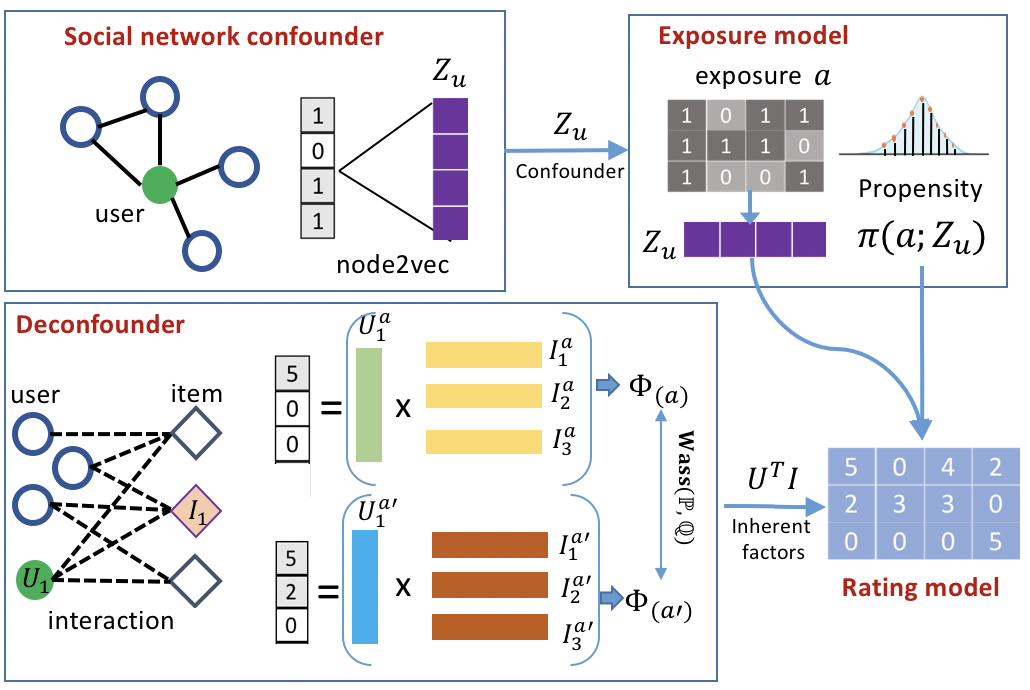}  
\caption{Our DENC method consists of four components: \emph{Social network confounder}, \emph{exposure model}, \emph{deconfonder model} and \emph{rating model}.}
\label{fig:frame}
\end{figure}

Followed the causal graph in Figure~\ref{fig:underlying}, we now design our DENC method incorporates three determinants in Figure~\ref{fig:frame}.
Each component accordingly corresponds to one of three specific determinants: \emph{social network confounder}, \emph{exposure model} and \emph{deconfonder model}, which jointly determine the rating outcome.

\subsection{Exposure Model}
To cope with the selection bias caused by users or the external social relations, we build on the causal inference theory and propose an effective exposure model. Guided by the treatment assignment mechanism in causal inference, we propose a novel exposure model that computes the probability of exposure variable specific to the user-item pair. This model is beneficial to understand the generation of the \emph{Missing Not At Random} (MNAR) patterns in ratings, which thus remedies selection biases in a principled manner. 
For example, user A goes to watch the movie because of his friend's strong recommendation. Thus, we propose to mitigate the selection bias by exploiting the network connectivity information that indicating \emph{to which extent the exposure for a user will be affected by its neighbors}. 

\subsubsection{Social Network Confounder}
To control the selection bias arisen from the external social network, we propose a confounder representation model that quantifies the common biased factors affecting both the exposure and rating outcome.

We now discuss the method of choosing and learning exposure. Let $G$ present the social relationships among users $U$, where an edge denotes there is a friend relationship between users. 
We resort to node2vec~\cite{grover2016node2vec} method and learn network embedding from diverse connectivity provided by the social network. 
More details about node2vec method can be found in Section~\ref{sec:embedding} in the appendix. 
To mine the deep social structure from $G$, for every source user $u$, node2vec generates the network neighborhoods $N_s(u) \subset G$ of node $u$ through a sampling strategy to explore its neighborhoods in a breadth-first sampling as well as a depth-first sampling manner. 
The representation $Z_u$ for user $u$ can be learned by minimizing the negative likelihood of preserving network neighborhoods $N_s(u)$: 
\begin{equation}
\begin{split}
    \mathcal{L}_{z}&=-{\sum_{u \in G}{\log P(N_{s}(u)|Z_u)}}\\
    &=\sum_{u \in G}\left[\log \sum_{v \in G} \exp (Z_v\cdot Z_u)-\sum_{u_{i} \in N_{s}(u)} Z_{u_{i}}\cdot Z_u\right]
    \label{eq:l_z}
\end{split}
\end{equation}
The final output $Z_u\in \mathbb{R}^d$ sufficiently explores diverse neighborhoods of each user, which thus represents to what extent the exposure for a user is influenced by his friends in graph $G$.

\subsubsection{Exposure Assignment Learning} 
The exposure under the recommendation scenario is not randomly assigned. Users in social networks often express their own preferences over the social network, which therefore will affect their friends' exposure policies. In this section, to characterize the \emph{Missing Not At Random} (MNAR) pattern in ratings, we resort to causal inference~\cite{pearl2009causality} to build the exposure mechanism influenced by social networks.

To begin with, we are interested in the binary exposure $a_{ui}$ that defines whether the item $i$ is exposed ($a_{ui}=1$) or unexposed ($a_{ui}=1$) to user $u$, i.e., $a_{ui}=1$. 
Based on the informative confounder learned from social network, we propose the notation of \emph{propensity} to capture the exposure from the causal inference language.
\begin{definition}[Propensity]
\label{df:pro}
Given an observed rating $y_{ui}\in\text{rating}$ and confounder $Z_u$ in~\eqref{eq:l_z}, the propensity of the corresponding exposure for user–item pair $(u,i)$ is defined as
\begin{equation}
    \pi(a_{ui};Z_u)=P(a_{ui}=1|y_{ui}\in\text{rating};Z_u)
    \label{eq:prp}
\end{equation}
\end{definition}
In view of the foregoing, 
we model the exposure mechanism by the probability of $a_{ui}$ being assigned to 0 or 1.
\begin{equation}
    \begin{aligned} P(a_{ui}) &=\prod_{u,i} P\left(a_{ui} \right) =\prod_{(u,i)\in\mathcal{O}} P\left(a_{ui}=1 \right) \prod_{(u,i)\notin\mathcal{O}} P\left(a_{ui}=? \right) \end{aligned}
    \label{eq:pa_all}
\end{equation}
where $\mathcal{O}$ is an index set for the observed ratings.
The case of $a_{ui}=1$ can result in an observed rating or unobserved rating: 1) for the observed rating represented by $y_{ui}\in \text{rating}$, we definitely know the item $i$ is exposed, i.e., $a_{ui}=1$; 2) an unobserved rating $y_{ui}\notin \text{rating}$ may represent a negative feedback (i.e., the user is not reluctant to rating the item) on the exposed item $a_{ui}=1$. In light of this, based on~\eqref{eq:prp}, we have
\begin{equation}
\begin{split}
    P(a_{ui}=1)&=P(a_{ui}=1, y_{ui}\in\text{rating})+P(a_{ui}=1,y_{ui}\notin\text{rating})\\
    &=\pi(a_{ui};Z_u)P(y_{ui}\in\text{rating})+W_{ui}P(y_{ui}\notin\text{rating})
\end{split}
\label{eq:pa_1}
\end{equation}
where $W_{ui}=P(a_{ui}=1|y_{ui}\notin \text{rating})$. The exposure $a_{ui}$ that is unknown follows the distributions as
\begin{equation}
\begin{split}
    P(a_{ui}=?)=1- P(a_{ui}=1)
\end{split}
\label{eq:pa_u}
\end{equation}

By substituting Eq.~\eqref{eq:pa_1} and Eq.~\eqref{eq:pa_u} for Eq.~\eqref{eq:pa_all}, we attain the exposure assignment for the overall rating data as
\begin{equation}
    P(a_{ui})=\prod_{(u, i) \in \mathcal{O}} \pi(a_{ui};Z_u)
    \prod_{(u, i)\notin \mathcal{O}} \left(1-W_{ui}\right)
    \label{eq:obj_p}
\end{equation}
Inspired by~\cite{pan2008one}, we assume uniform scheme for $W_{ui}$ when no side information is available. 
According to most causal inference methods~\cite{shalit2017estimating,pearl2009causality}, a widely-adopted parameterization for $\pi(a_{ui};Z_u)$ is a logistic regression network parameterized by $\Theta=\{W_0,b_{0}\}$, i.e.,
\begin{equation}
  \pi(a_{ui};Z_u,\Theta)=\mathbb{I}_{y\in\text{rating}}\cdot\left[1+e^{-(2 a_{ui}-1)\left(Z_u^{\top}  \cdot W_0+b_{0}\right)}\right]^{-1} 
  \label{eq:pa}
\end{equation}
Based on Eq.~\eqref{eq:pa}, the overall exposure $P(a_{ui})$ in Eq.~\eqref{eq:obj_p} can be written as the function of parameters $\Theta=\{W_0,b_{0}\}$ and $Z_u$, i.e.,
\begin{equation}
    \begin{split}
        \mathcal{L}_{a}=\sum_{u,i}-\log P(a_{ui} ;Z_u, \Theta)
    \end{split}
    \label{eq:l_a}
\end{equation}
where social network confounder $Z_u$ is learned by the pre-trained node2vec algorithm. Similar to supervised learning, $\Theta$
can be optimized through minimization of the negative log-likelihood.

\subsection{Deconfounder Model}
Traditional recommendation learns the latent factor representations for user and item by minimizing errors on the observed ratings, e.g., matrix factorization. Due to the existence of selection bias, such a learned representation may not necessarily minimize the errors on the unobserved rating prediction.
Inspired by~\cite{shalit2017estimating}, we propose to learn a balanced representation that is independent of exposure assignment such that it represents inherent or invariant features in terms of users and items. 
The invariant features must also lie in the inference space shown in Figure~\ref{fig:seb}, which can be used to consistently infer unknown ratings using observed ratings.
This makes sense in theory: if the learned representation is hard to distinguish across different exposure settings, it represents invariant features related to users and items.

According to Figure~\ref{fig:underlying}, we can define two latent vectors $U\in\mathbb{R}^{k_d}$ and $I\in\mathbb{R}^{k_d}$ to represent the inherent factor of a user and a item, respectively.
Recall that different values for $W_{ui}$ in Eq.~\eqref{eq:obj_p} can generate different exposure assignments for the observed rating data. Following this intuition, we construct two different exposure assignments $a$ and $\hat{a}$ corresponding two settings of $W_{ui}$. Accordingly, $\Phi_{(a)}$ and $\Phi_{(\hat{a})}$ are defined to include inherent factors of users and items, i.e., $\Phi_{(a)}=\left[U_{1}^{(a)},\cdots, U_{M}^{(a)},  I_{1}^{(a)},\cdots,I_{M}^{(a)}\right] \in \mathbb{R}^{k_d\times 2M }$, $\Phi_{(\hat{a})}=\left[U_{1}^{(\hat{a})},\cdots, U_{M}^{(\hat{a})},  I_{1}^{(\hat{a})},\cdots,I_{M}^{(\hat{a})}\right] \in \mathbb{R}^{k_d \times 2M}$. 
Figure~\ref{fig:underlying} also indicates that the inherent factors of user and item would keep unchanged even if the exposure variable is altered from 0 to 1, and vice versa. 
That means $U\in\mathbb{R}^{k_d}$ and $I\in\mathbb{R}^{k_d}$ should be independent of the exposure assignment, i.e., $U^{(a)}\indep U^{(\hat{a})}$ or $I^{(a)}\indep I^{(\hat{a})}$.
Accordingly, minimizing the discrepancy between $\Phi_{(a)}$ and $\Phi_{(\hat{a})}$ ensures that the learned factors embeds no information about the exposure variable and thus reduce selection bias. The penalty term for such a discrepancy is defined as
\begin{equation} 
\begin{split}
   \mathcal{L}_{d}=\text{disc}\left(\Phi_{(\hat{a})},\Phi_{(a)}\right)\\
\end{split}
\label{eq:la}
\end{equation}

Inspired by~\cite{muller1997integral}, we employ \emph{Integral Probability Metric} (IPM) to estimate the discrepancy between $\Phi_{(\hat{a})}$ and $\Phi_{(a)}$.  $\text{IPM}_{\mathcal{F}}(\cdot,\cdot)$ is the (empirical) integral probability metric defined by the function family $\mathcal{F}$. 
Define two probability distributions $\mathbb{P}=P(\Phi_{(\hat{a})})$ and $\mathbb{Q}=P(\Phi_{(a)})$, the corresponding IPM is denoted as
\begin{equation}
    \text{IPM}_{\mathcal{F}}(\mathbb{P},\mathbb{Q})=\sup _{f \in \mathcal{F}}\left|\int_{S} f d \mathbb{P}-\int_{S} f d \mathbb{Q}\right|
\end{equation}
where $\mathcal{F}:S\rightarrow \mathbb{R}$ is a class of real-valued bounded measurable functions. 
We adopt $\mathcal{F}$ as 1-Lipschitz functions that lead IPM to the Wasserstein-1 distance, i.e.,
\begin{equation}
    Wass(\mathbb{P},\mathbb{Q})=\inf _{f \in \mathcal{F}} \sum_{\mathbf{v}\in\text{col}_{i}(\Phi_{(\hat{a})})}\|f(\mathbf{v})-\mathbf{v}\| \mathbb{P}(\mathbf{v}) d \mathbf{v}
    \label{eq:l_ind}
\end{equation}
where $\mathbf{v}$ is the $i$-th column of $\Phi_{(\hat{a})}$ and the set of push-forward functions $\mathcal{F}=\left\{f \mid f: \mathbb{R}^{d} \rightarrow \mathbb{R}^{d} \text { s.t. } \mathbb{Q}(f(\mathbf{v}))=\mathbb{P}(\mathbf{v})\right\}$ can transform the representation distribution of the exposed $\Phi_{(\hat{a})}$ to that of the unexposed $\Phi_{(a)}$. Thus, $\|f(\mathbf{v})-\mathbf{v}\|$ is a pairwise distance matrix between the exposed and unexposed user-item pairs.
Based on the discrepancy defined in~\eqref{eq:l_ind}, we define $C(\Phi)=\|f(\mathbf{v})-\mathbf{v}\|$ and reformulate penalty term in~\eqref{eq:la} as 
\begin{equation}
    \mathcal{L}_{d}=\inf_{\gamma \in \Pi\left(\mathbb{P}, \mathbb{Q}\right)} \mathbb{E}_{(\mathbf{v}, f(\mathbf{v})) \sim \gamma}
    C(\Phi)
    \label{eq:l_ind}
\end{equation}

We adopt the efficient approximation algorithm proposed by~\cite{shalit2017estimating} to compute the gradient of~\eqref{eq:l_ind} for training the deconfounder model.
In particular, a mini-batch with $l$ exposed and $l$ unexposed user-item pairs is sampled from $\Phi_{(\hat{a})}$ and $\Phi_{(a)}$, respectively. The element of distance matrix $C(\Phi)$ is calculated as $C_{ij}=\|\text{col}_{i}(\Phi_{(\hat{a})})-\text{col}_{j}(\Phi_{(a)})\|$. 
After computing $C(\Phi)$, we can approximate $f$ and the gradient against the model parameters
~\footnote{For a more detailed calculation, refer to Algorithm 2 in the appendix of prior work~\cite{shalit2017estimating}}.
In conclusion, the learned latent factors generated by the deconfounder model embed no information about exposure variable. That means all the confounding factors are retained in social network confounder $Z_u$.

\subsection{Learning}
\subsubsection{Rating prediction}
Having obtained the final representations $U$ and $I$ by the deconfounder model, we use an inner product of $U^{\top}I$ as the inherent factors to estimate the rating. 
As shown in the causal structure in Figure~\ref{fig:frame}, another component affecting the rating prediction is the social network confounder. 
A simple way to incorporate these components into recommender systems is through a linear model as follows.
\begin{equation}
    \hat{y}_{ui}=\sum_{u,i\in \mathcal{O}}   U^{\top}I+ {W_u}^{\top} Z_u+\epsilon_{ui},\quad \epsilon_{ui}\sim\mathcal{N}(0,1)
\end{equation}
where $W_u$ is coefficient that describes how much the confounder $Z_u$ contributes to the rating.
To define the unbiased loss function for the biased observations $y_{ui}$, we leverage the IPS strategy~\cite{schnabel2016recommendations} to weight each observation with \emph{Propensity}. By  Definition~\ref{df:pro}, the intuition of the inverse propensity is to down-weight the commonly observed ratings while up-weighting the rare ones.
\begin{equation}
    \mathcal{L}_{y}=\frac{1}{|\mathcal{O}|}\sum_{u,i\in \mathcal{O}}\frac{ \left(y_{ui}-\hat{y}_{ui}\right)^{2}}{\pi(a_{ui};Z_u)}
    \label{eq:loss_y}
\end{equation}

\subsubsection{Optimization}

To this end, the objective function of our DENC method to predict ratings could be derived as:
\begin{equation}
\begin{aligned}
\mathcal{L}= \mathcal{L}_{y} +\lambda_a\mathcal{L}_{a}+\lambda_z\mathcal{L}_{z}+ \lambda_d\mathcal{L}_{d}+\mathcal{R}(\Omega)
\end{aligned} 
\end{equation}
where $\Omega$ represents the trainable parameters and $\mathcal{R}(\cdot)$ is a squared $l_2$ norm regularization term on $\Omega$ to alleviate the overfitting problem. $\lambda_a$, $\lambda_z$ and $\lambda_d$ are trade-off hyper-parameters.
To optimize the objective function, we adopt
Stochastic Gradient Descent(SGD)~\cite{bottou2010large}
as the optimizer due to its efficiency.

\section{Experiments}
To more thoroughly understand the nature of MNAR issue and the proposed unbiased DENC, 
experiments are conducted to answer the following research questions:
\begin{itemize}[leftmargin=*]
    \item (\textbf{RQ1}) How confounder bias caused by the social network is manifested in real-world recommendation datasets?
    \item (\textbf{RQ2}) Does our DENC method achieve the state-of-the-art performance in debiasing recommendation task?
    \item (\textbf{RQ3}) How does the embedding size of each component (e.g., social network confounder and deconfounder model) in our DENC method impact the debiasing performance?
    \item (\textbf{RQ4}) How do the missing social relations impact the debiasing performance of our DENC method?
\end{itemize}

\subsection{Setup}
\subsubsection{Evaluation Metrics}
We adopt two popular metrics including Mean Absolute Error (MAE) and Root Mean Square Error (RMSE) to evaluate the performance.
Since improvements in MAE or RMSE terms can have a significant impact on the quality of the Top-$K$ recommendations~\cite{koren2008factorization}, we also evaluate our DENC with Precision@K and Recall@K for the ranking performance\footnote{We consider items with a rating greater than or equal to 3.5 as relevant}. 
\subsubsection{Datasets}\label{dataset}

We conduct experiments on three datasets including one semi-synthetic dataset and two benchmark datasets \texttt{Epinions} \footnote{http://www.cse.msu.edu/~tangjili/trust.html} and \texttt{Ciao}~\cite{tang-etal12a} \footnote{http://www.cse.msu.edu/~tangjili/trust.html}. 
We maintain all the user-item interaction records in the original datasets instead of discard items that have sparse interactions with users.\footnote{Models can benefit from the preprocessed datasets in which all items interact with at least a certain amount of users, for such preprocessing will reduce the dataset sparsity. }
The semi-synthetic dataset is generated by incorporating the social network into \texttt{MovieLens}\footnote{https://grouplens.org/datasets/movielens} dataset. The details of these datasets are given in Section~\ref{sec:data} in the appendix.

\subsubsection{Baselines}
We compare our DENC against three groups of methods for rating prediction: (1) \textbf{Traditional methods}, including NRT~\cite{li2017neural} and PMF~\cite{mnih2008probabilistic}. (2) \textbf{Social network-based methods}, including GraphRec~\cite{fan2019graph}, DeepFM+~\cite{guo1703factorization}, SocialMF~\cite{jamali2010matrix}, SREE~\cite{li2017social} and SoReg~\cite{ma2011recommender}.
(3) \textbf{Propensity-based methods}, including CausE~\cite{bonner2018causal} and D-WMF~\cite{wang2018deconfounded}. More implementation details of baselines and parameter settings are included in Section~\ref{sec:baseline} in the appendix.

\begin{table}[!htb]
  \caption{Statistics of Datasets. Density for rating (density-R) is $\#ratings/(\#users \cdot\#items)$, Density for social relations (density-SR) is $\#relations/(\#users \cdot\#users)$.}
  \label{tab:2}
  \resizebox{0.4\textwidth}{!}{
  \begin{tabular}{cccc}
    \toprule
     &\texttt{Epinions} & \texttt{Ciao} & \texttt{MovieLens-1M}\\
    \midrule
    \# \texttt{users}& 22,164& 7,317& 6,040\\
    \# \texttt{items} &296,277 & 104,975 &3,706 \\\hline
    \# \texttt{ratings}& 922,267 &283,319 &1000,209\\
    \texttt{density-R} (\%) & 0.0140 & 0.0368 &4.4683\\
    \hline
    \# \texttt{relations} & 355,754&111,781 & 9,606\\
    \texttt{density-SR} (\%) & 0.0724 &0.2087 & 0.0263\\
    \bottomrule
  \end{tabular}
  }
\end{table}
\subsubsection{Parameter Settings}
We implement all baseline models on a Linux server with Tesla P100 PCI-E 16GB GPU.~\footnote{Our code is currently shared on Github, however, due to the double-blind submission policy requirement, we leave the link void now but promise to activate it after paper acceptance.}
Datasets for all models except CausE~\footnote{
As in CausE, we sample 10\% of the training set to build an additional debiased dataset (mandatory in model training),
where items are sampled to be uniformly exposed to users.
} 
are split as training/test sets with a proportion of 80/20, and 20\% of the training set are validation set.

We optimize all models with Stochastic Gradient Descent(SGD)~\cite{bottou2010large}.
For fair comparison, a grid search is conducted to choose the optimal parameter settings, e.g., 
dimension of user/item latent vector 
$k_{MF}$ for matrix factorization-based models 
and dimension of embedding vector $d$ for neural network-based models. The embedding size is initialized with the Xavier~\cite{glorot2010understanding} and searched in $[8, 16, 32, 64, 128, 256]$. The batch size and learning rate are searched in $[32, 64, 128, 512, 1024]$ and $[0.0005,
0.001, 0.005, 0.01, 0.05, 0.1]$, respectively.
The maximum epoch $N_{epoch}$ is set as 2000, an early stopping strategy is performed.
Moreover,
we employ three hidden layers for the neural components of NRT, GraphRec and DeepFM+.
Like our DENC method, DeepFM+ uses node2vec to train the social network embeddings.
Hence, the embedding size of its node2vec is set as the same as in our DENC for a fair comparison.

Without specification, unique hyperparameters of DENC are
set as: 
three coefficients $\lambda_n$, $\lambda_u$ and $\lambda_i$ are tuned in $\{10^{-3},10^{-4},10^{-5}\}$.
The dimension of node2vec embedding size $k_a$ and the dimension of inherent factor $k_{d}$ are tuned in $[ 8, 16, 32, 64, 128, 256 ]$, and their influences are reported in Section~\ref{ablation}.

\begin{table*}
\centering
  \caption{
  Performance comparison: bold numbers are the best results. Strongest baselines are highlighted with underlines.
  }
  \label{tab:4}
\resizebox{0.8\textwidth}{!}{%
\begin{tabular}{c||c||cc||ccccc||cc||c}
\hline
 &  & \multicolumn{2}{c||}{\textbf{Traditional}} & \multicolumn{5}{c||}{\textbf{Social network-based}} & \multicolumn{2}{c||}{\textbf{Propensity-based}} & \textbf{Ours} 
 \\ [3pt]\hline 
\hline
    Dataset & Metrics & PMF & NRT & SocialMF & SoReg & SREE & GraphRec & DeepFM+ & CausE & D-WMF& \textbf{DENC}\\\hhline{-||-||--||-----||--||-}
    \multicolumn{1}{c||}{\texttt{Epinions}} & MAE & 0.9505 & 0.9294 & 0.8722 & 0.8851 & 0.8193 & 0.7309 & 0.5782 & 0.5321 & \underline{0.3710} &\textbf{0.2684}  \\[3pt]
\multicolumn{1}{c||}{} & RMSE & 1.2169 & 1.1934 & 1.1655 & 1.1775 & 1.1247 & 0.9394 & 0.6728 & 0.7352 & \underline{0.6299} & \textbf{0.5826}  \\ [3pt]\hline
\multicolumn{1}{c||}{\texttt{Ciao}} & MAE & 0.8868 & 0.8444 & 0.7614 & 0.7784 & 0.7286 &0.6972 & 0.3641 & 0.4209 & \underline{0.2808} &\textbf{0.2487}  \\[3pt]
\multicolumn{1}{c||}{} & RMSE &1.1501 & 1.1495 & 1.0151 &1.0167 & 0.9690 &0.9021 & 0.5886 & 0.8850 & \underline{0.5822} &\textbf{0.5592}  \\[3pt] \hline
\multicolumn{1}{c||}{\texttt{MovieLens-1M}} & MAE & 0.8551 & 0.8959 & 0.8674 & 0.9255 & 0.8408 & 0.7727 & 0.5786 &0.4683 & \underline{0.3751} & \textbf{0.2972}  \\[3pt]
\multicolumn{1}{c||}{$\Delta(Z_u)=-0.35$} & RMSE & 1.0894 &1.1603 & 1.1161 & 1.1916 & 1.0748 & 0.9582 & 0.6730  &0.8920 & \underline{0.6387} &\textbf{0.5263}  \\[3pt] \hline
\multicolumn{1}{c||}{\texttt{MovieLens-1M}} & MAE & 0.8086 & 0.8801 & 0.8182 & 0.8599 & 0.7737 & 0.7539 & 0.5281 & 0.4221 & \underline{0.3562}  & \textbf{0.2883}  \\[3pt]
\multicolumn{1}{c||}{$\Delta(Z_u)=0$} & RMSE & 1.0034 & 1.1518 & 1.0382 & 1.1005 & 0.9772 & 0.9454 & 0.6477 & 0.8333 & \underline{0.6152} &\textbf{0.5560}  \\ [3pt]\hline
\multicolumn{1}{c||}{\texttt{MovieLens-1M}} & MAE & 0.7789 &0.7771 & 0.7969 & 0.8428 & 0.7657 & 0.7423 &0.3672 & 0.4042 & \underline{0.3151} & \textbf{0.2836}  \\[3pt]
\multicolumn{1}{c||}{$\Delta(Z_u)=0.35$} & RMSE & 0.9854 &0.9779 & 1.0115 & 1.0792 & 0.9746 & 0.9344 &\underline{0.5854} & 0.8173 & 0.5962 &\textbf{0.5342}  \\[3pt] \hline
\end{tabular}
}
\end{table*}

\subsection{Understanding Social Confounder (RQ1)}
We initially conduct an experiment to understand to what extent the confounding bias caused by social networks is manifested in real-world recommendation datasets.
The social network as a confounder will bias the interactions between the user and items.
We aim to verify two kinds of scenarios: (1) User in the social network interacts with more items than users outside the social network. (2) The pair of user-neighbor in the social network has more common interacted items than the pair of user-neighbor outside the social network.
Intuitively, an unbiased platform should expect users to interact with items broadly, which indicates that interactions are likely to be evenly distributed.
Thus, we investigate the social confounder bias by analyzing the statistics of interactions in these two scenarios in \texttt{Epinions} and \texttt{Ciao} dataset. 

\begin{figure}[htbp]
\centering
\begin{minipage}[t]{0.22\textwidth}
\centering
\includegraphics[width=\textwidth]{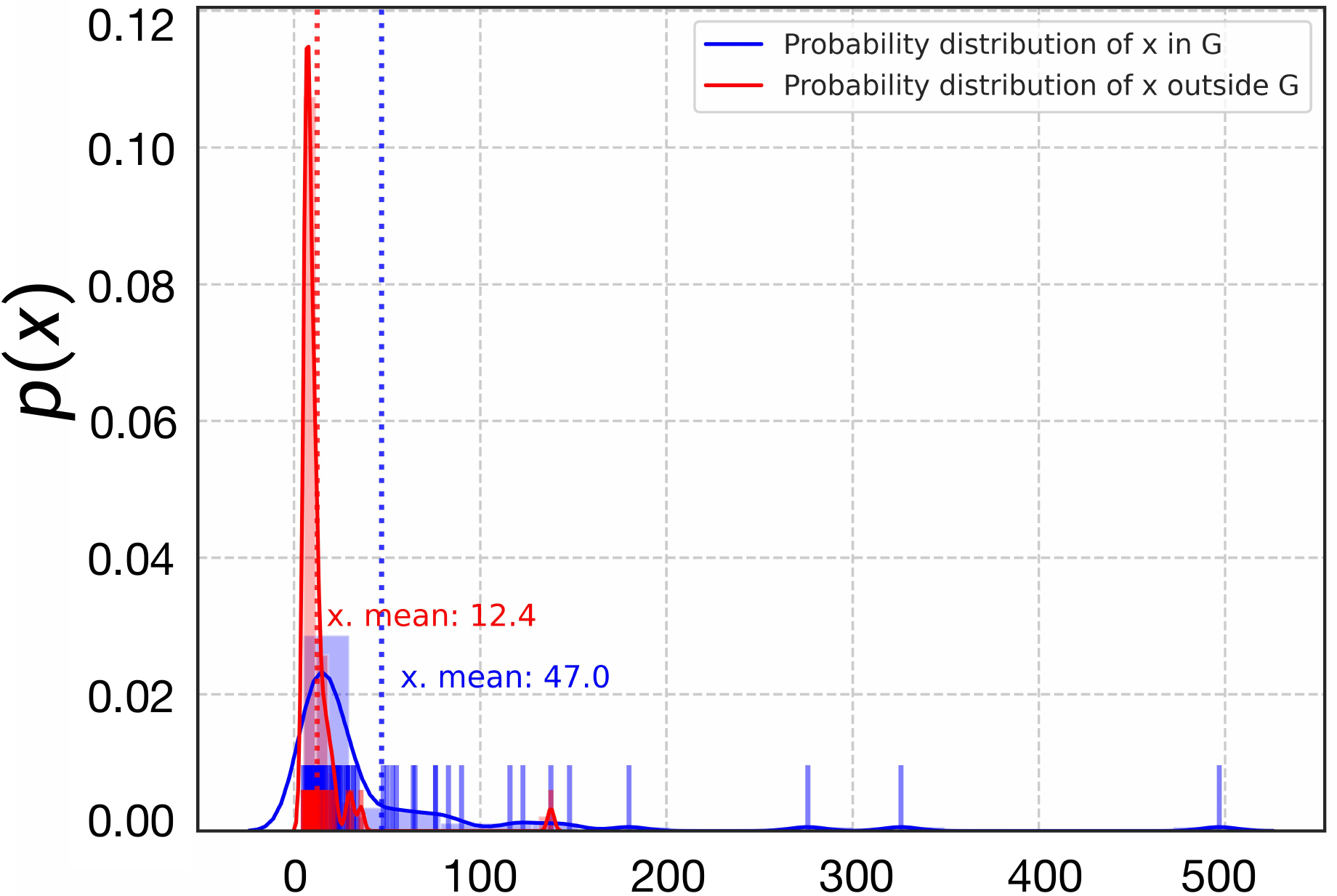}
\subcaption*{(a) Distribution on \texttt{Ciao}.}
  \label{fig:prove_1_ciao}
\end{minipage}
\begin{minipage}[t]{0.22\textwidth}
\centering
\includegraphics[width=\textwidth]{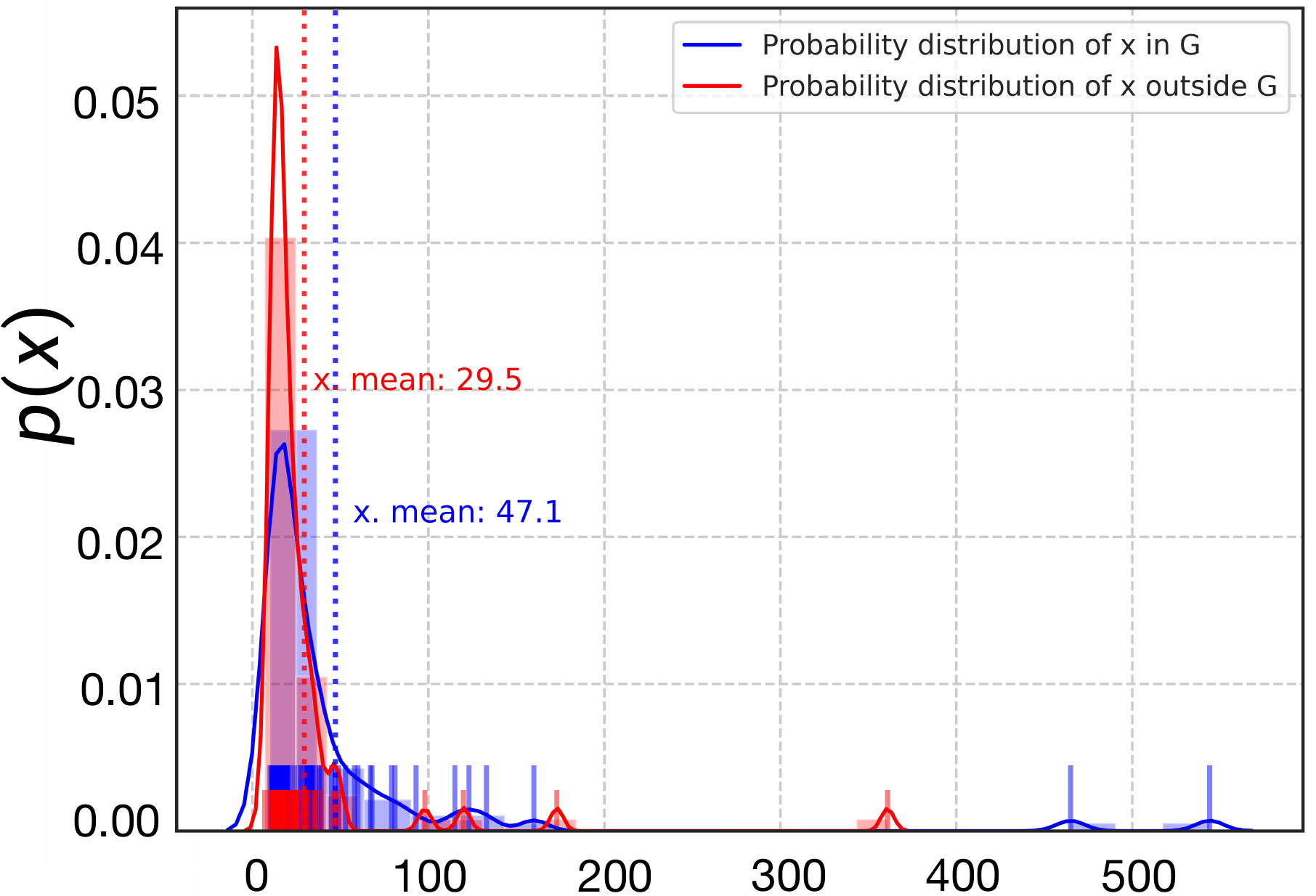}
\subcaption*{(b) Distribution on \texttt{Epinions}.}
  \label{fig:prove_1_epinion}
\end{minipage}
\caption{Scenario (1): the distribution of $x$ (the number of items interacted by a user). The smooth probability curves visualize how the number of items is distributed. 
}
\label{fig:prove_1}
\end{figure}

\begin{figure}[htbp]
\centering
\begin{minipage}[t]{0.22\textwidth}
\centering
\includegraphics[width=\textwidth]{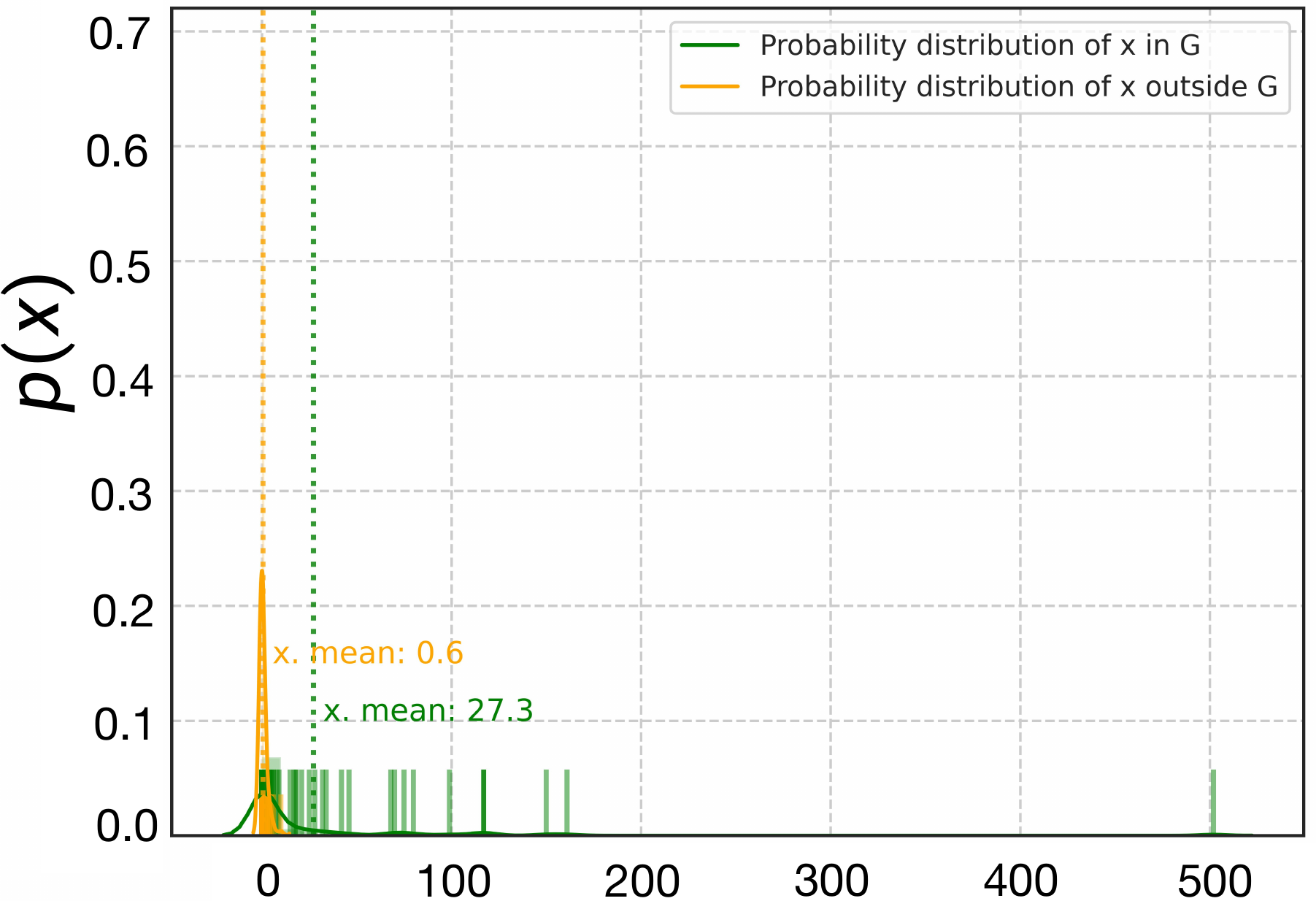}
\subcaption*{(a) Distribution on \texttt{Ciao}.}
  \label{fig:prove_2_ciao}
\end{minipage}
\begin{minipage}[t]{0.22\textwidth}
\centering
\includegraphics[width=\textwidth]{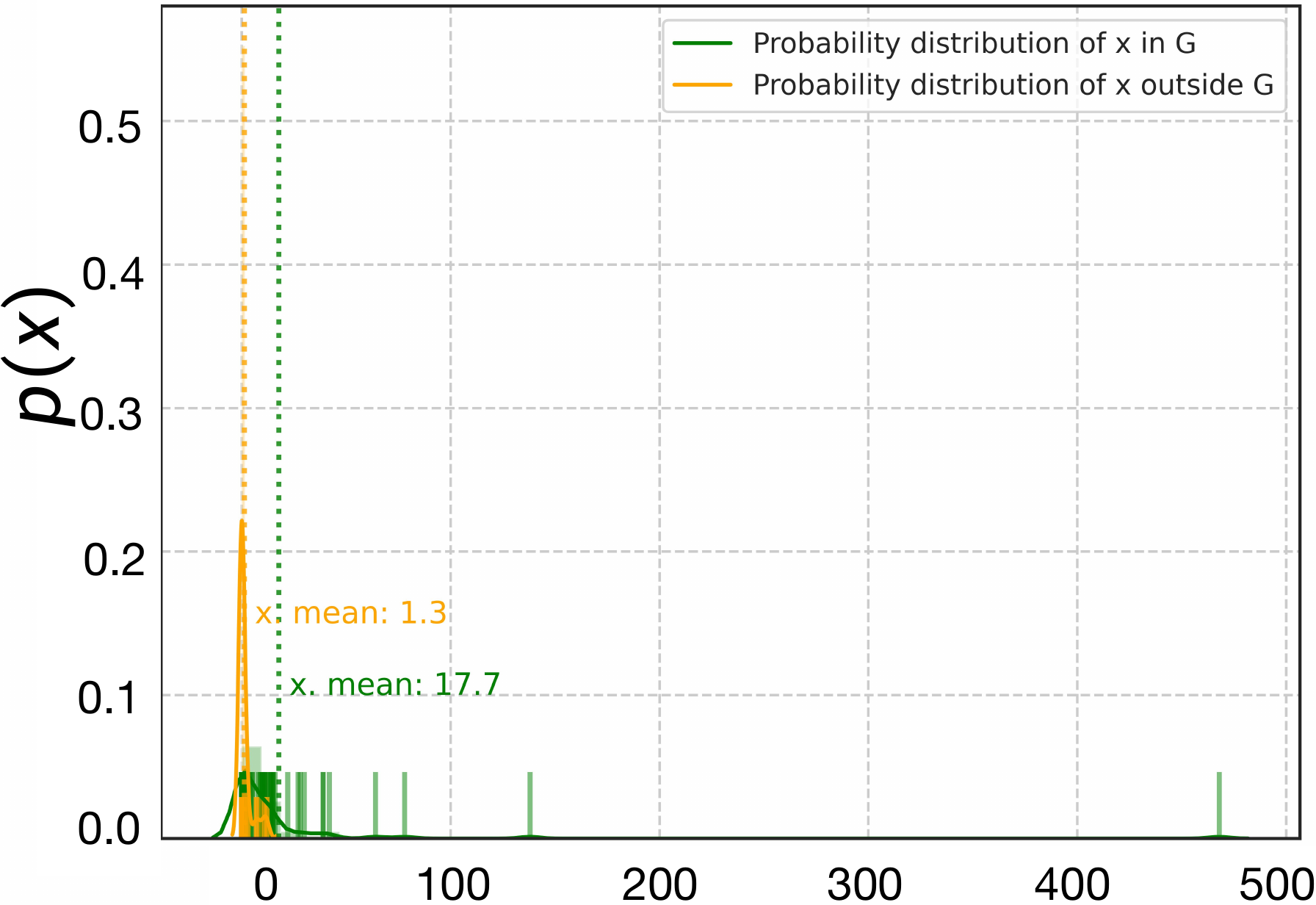}
\subcaption*{(b) Distribution on \texttt{Epinions}.}
  \label{fig:prove_2_epinion}
\end{minipage}
\caption{Scenario (2): the distribution of $x$ (the number of items commonly interacted by a user-pair). 
}
\label{fig:prove_2}
\end{figure}

For the first scenario, we construct two user sets within or outside the social network, i.e., $\mathcal{U}_{G}$ and $\mathcal{U}_{\backslash G}$. Specially, $\mathcal{U}_{G}$ is constructed by randomly sampling a set of users in social network $G$, and $\mathcal{U}_{\backslash G}$ is randomly sampled out of $G$. The size of $\mathcal{U}_{G}$ and $\mathcal{U}_{\backslash G}$ is the same and defined as $n$.
Following the above guidelines, we sample $n=70$ users for $\mathcal{U}_{G}$ and $\mathcal{U}_{\backslash G}$.
Figure~\ref{fig:prove_1} depicts the distributions of the interacted items by users in $\mathcal{U}_{G}$ and $\mathcal{U}_{\backslash G}$. The smooth curves are continuous distribution estimates produced by the kernel density estimation. 
Apparently, the distribution for $\mathcal{U}_{\backslash G}$ is significantly skewed: Most of the users interact with few items. For example, on \texttt{Ciao}, more than 90\% of users interact with fewer than 50 items.
By contrast, most users in the social network tend to interact with items more frequently, which is also confirmed by the even distribution. In general, the distribution curve of $\mathcal{U}_{G}$ is quite different from $\mathcal{U}_{\backslash G}$, which reflects that the social network influences the interactions between users and items. In addition, the degree of bias varies across datasets: \texttt{Epinions} is less biased than \texttt{Ciao}.

For the second scenario, based on $\mathcal{U}_{G}$ and $\mathcal{U}_{\backslash G}$, we further analyze the number of commonly interacted items by the user-pair. Particularly, we randomly sample four one-hop neighbours for each user in $\mathcal{U}_{G}$ to construct user-pairs. Since users in $\mathcal{U}_{\backslash G}$ have no neighbours, for each of them, we randomly select another four users\footnote{According to the statistics, we discover that 90\% of users have at least four one-hop neighbours in \texttt{Ciao} and \texttt{Epinions}} in $\mathcal{U}_{\backslash G}$ to construct four user-pairs.
Recall that $\mathcal{U}_{G}$ and $\mathcal{U}_{\backslash G}$ both have 70 users, then we totally have $4\times 70$ user-pairs for $\mathcal{U}_{\backslash G}$ and $\mathcal{U}_{\backslash G}$, respectively.
Figure~\ref{fig:prove_2} represents the distribution of how many items are commonly interacted by the users in each pair.\footnote{For example, $\{user1, user2, user3, user4\}$ are one-hop neighbours of $user5$. If the number of commonly items interacted by $user1$ and $user5$ is 3, then $x=3$ in the $x$-axis of Figure~\ref{fig:prove_2} is nonzero.}
Figure~\ref{fig:prove_2} indicates most user-neighbour pairs in the social network have fewer than 10 items in common. However the user-pairs outside the social network nearly have no items in common, i.e., less than 1.
We can conclude that social networks can encourage users to share more items with their neighbours, compared with users who are not connected by any social networks. 

\subsection{Performance Comparison (RQ2)}
\label{sec:comp}
We compare the rating prediction of DENC with nine recommendation baselines on three datasets including \texttt{Epinions}, \texttt{Ciao} and \texttt{MovieLens-1M}. Table~\ref{tab:4} demonstrates the performance comparison, where the confounder $\Delta{(Z_u)}$ in \texttt{MovieLens-1M} is assigned with three different settings, i.e., -0.35, 0 and 0.35.
Analyzing Table~\ref{tab:4}, we have the following observations.

\begin{itemize}[leftmargin=*]
\item 
Overall, our DENC consistently yields the best performance among all methods on five datasets. 
For instance, DENC improves over the best baseline model w.r.t. MAE/RMSE by 10.26/4.73\%, 3.21/2.3\%, and 7.79/11.24\% on \texttt{Epinions}, \texttt{Ciao} and \texttt{MovieLens-1M} ($\Delta{(Z_u)}$=-0.35)
datasets, respectively. 
The results indicate the effectiveness of DENC on
the task of rating prediction, 
which has adopted a principled causal inference way to
leverage both the inherent factors and auxiliary social network information for improving recommendation performance.
\item Among the three kinds of baselines, propensity-based methods serves as the strongest baselines in most cases. This justifies the effectiveness of exploring the missing pattern in rating data by estimating the propensity score, which offers better guidelines to identify the unobserved confounder effect from ratings. However, propensity-based methods perform worse than our DENC, as they ignore the social network information. 
It is reasonable that exploiting the social network is useful to alleviate the confounder bias to rating outcome.
The importance of social networks can be further verified by the fact that most of the social network-based methods consistently outperform PMF on all datasets. 
\item All baseline methods perform better on \texttt{Ciao} than on \texttt{Epinions},  because \texttt{Epinions} is significantly sparser than \texttt{Ciao} with 0.0140\% and 0.0368\% density of ratings.
Besides this, DENC still achieves satisfying performance on \texttt{Epinions} and its performance is competitive with the counterparts on \texttt{Ciao}.
This demonstrates that its exposure model of DENC has an outstanding capability of identifying the missing pattern in rating prediction,
in which biased user-item pairs in \texttt{Epinions} can be captured and then alleviated.
In addition, the performance of DENC on three \texttt{Movielens-1M} datasets is stable w.r.t. different levels of confounder bias, which verifies the robust debiasing capability of DENC. 
\end{itemize}

\subsection{Ablation Study (RQ3)}\label{ablation}

In this section, we conduct experiments to evaluate the parameter sensitivity of our DENC method.  
We have two important hyperparameters $k_a$ and $k_{d}$ that correspond to the embedding size in loss function $\mathcal{L}_a$ and $\mathcal{L}_d$, respectively. 
Based on the hyperparameter setup in Section 4.1.4, we vary the value of one hyperparameter while keeping the others unchanged. 

\begin{figure}[htbp]
\centering
\begin{minipage}[t]{0.22\textwidth}
\centering
\includegraphics[width=\textwidth]{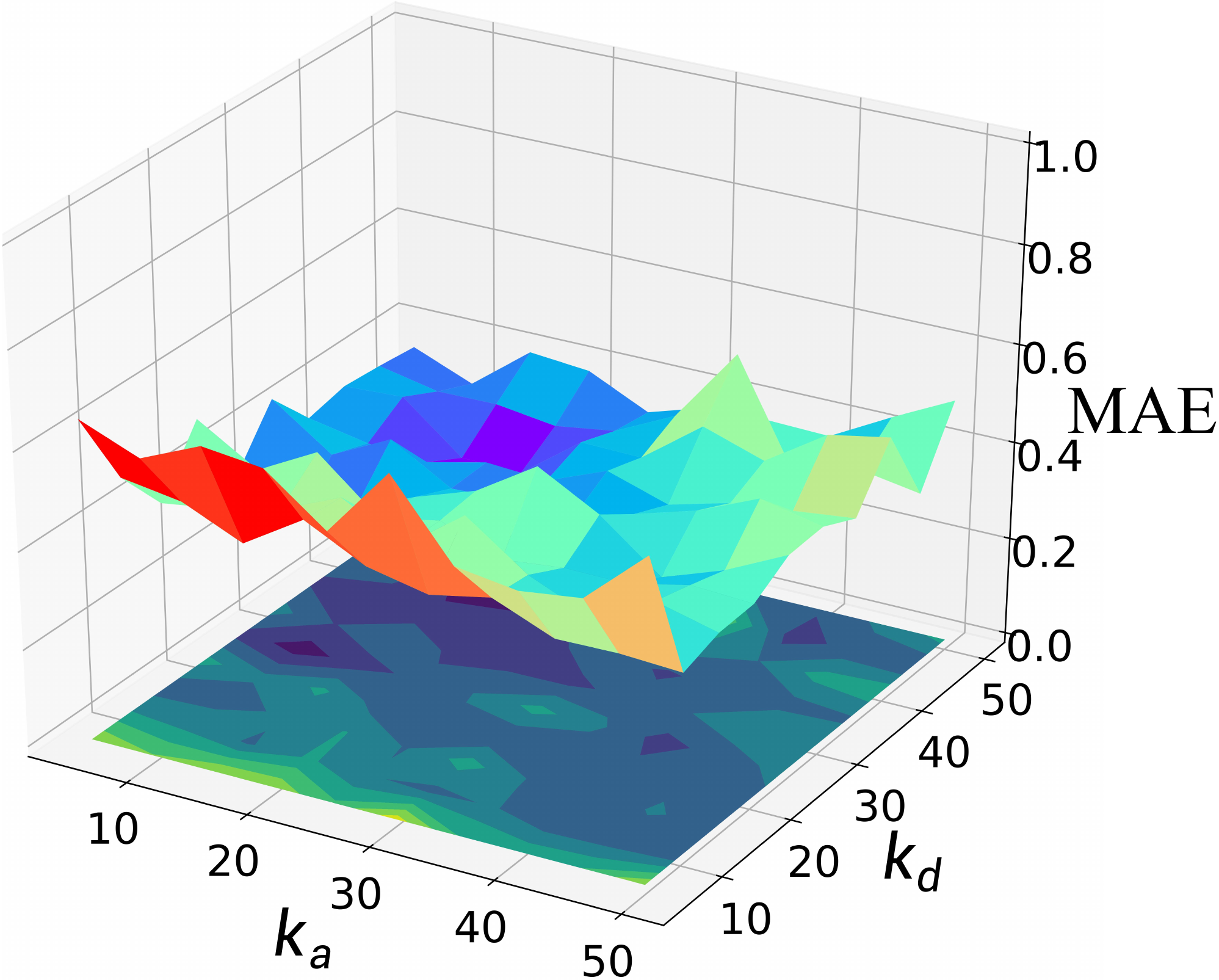}
\subcaption*{(a) MAE on Ciao.}
\end{minipage}
\begin{minipage}[t]{0.22\textwidth}
\centering
\includegraphics[width=\textwidth]{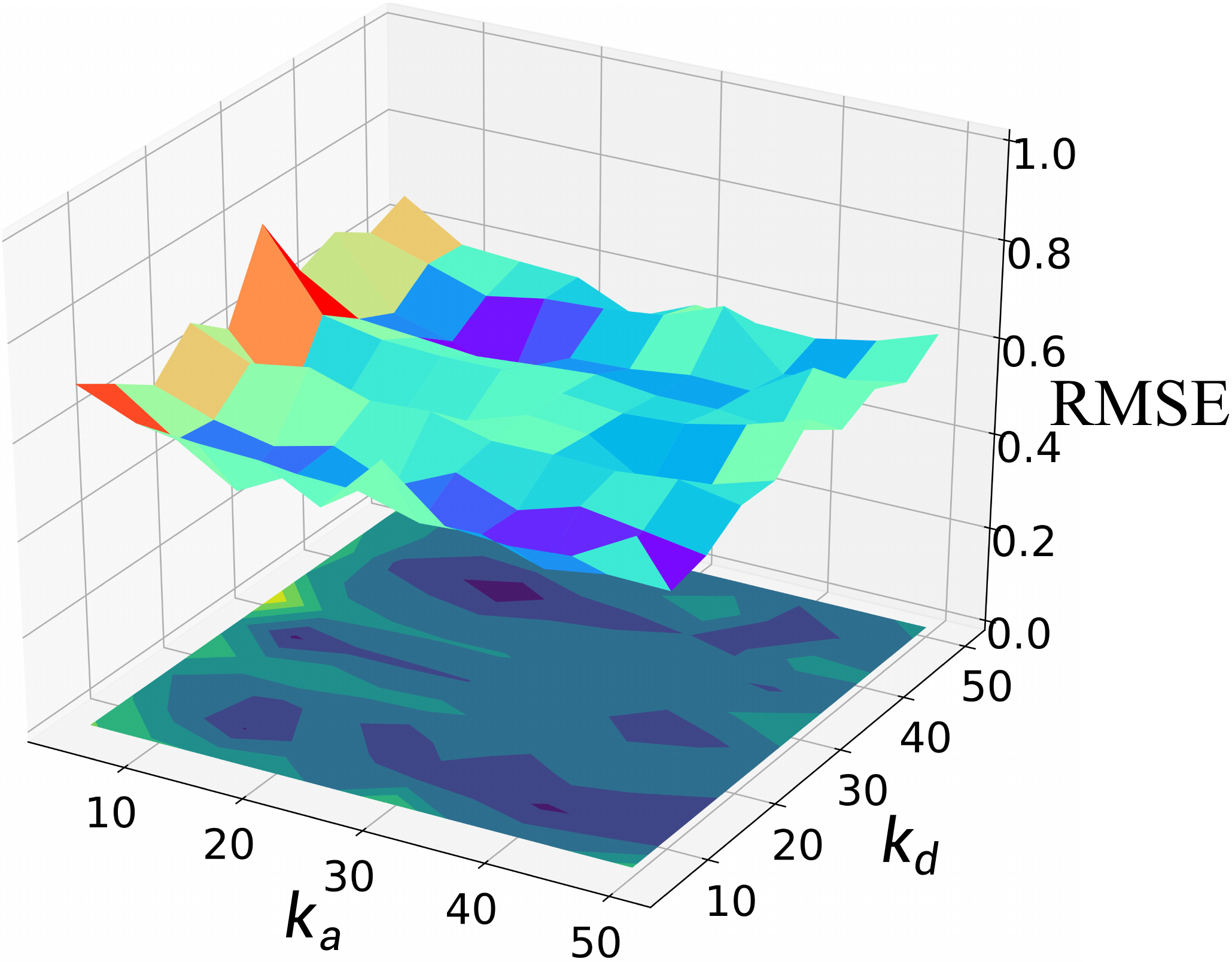}
\subcaption*{(b) RMSE on Ciao.}
\end{minipage}
\begin{minipage}[t]{0.22\textwidth}
\centering
\includegraphics[width=\textwidth]{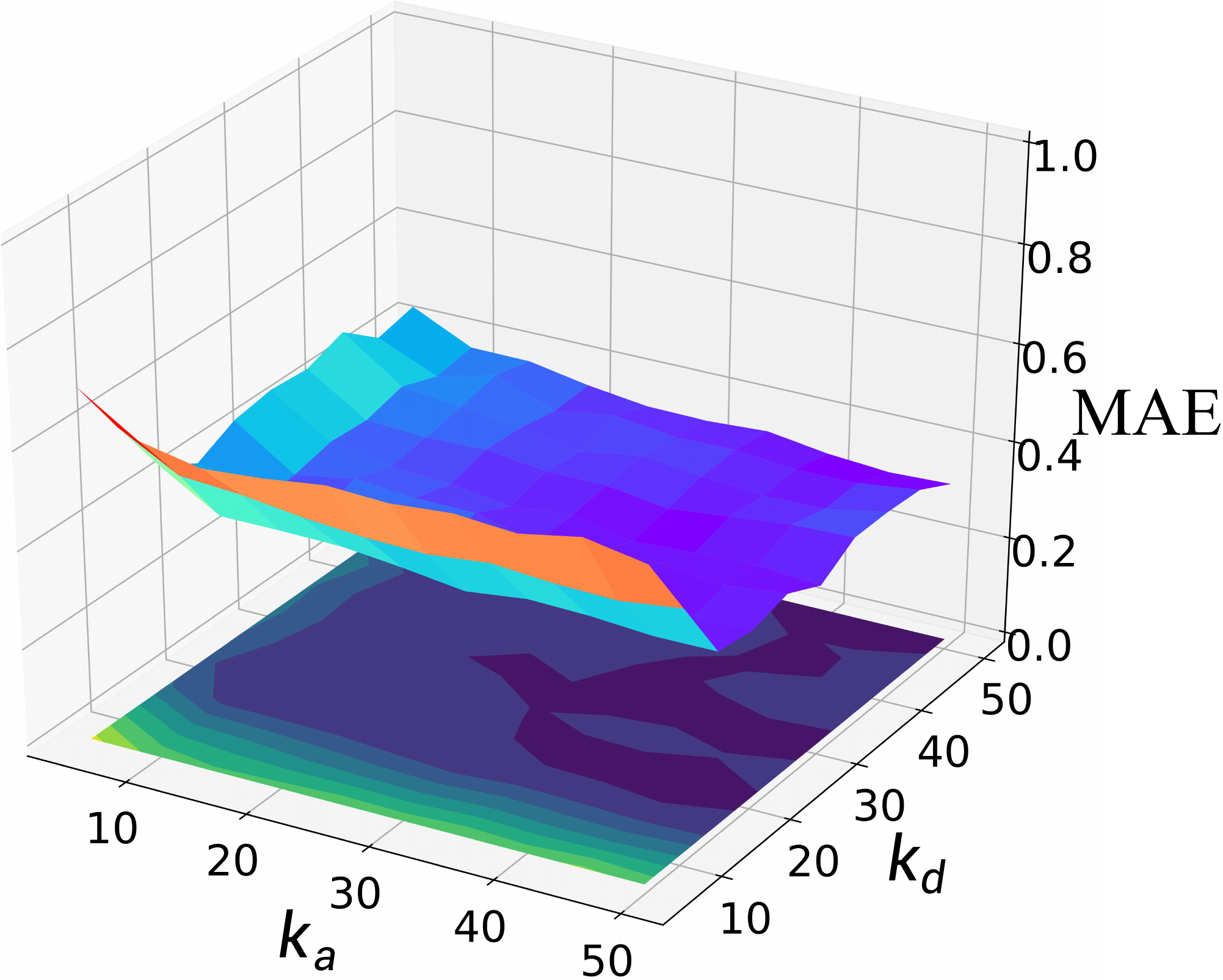}
\subcaption*{(a) MAE on Epinions.}
\end{minipage}
\begin{minipage}[t]{0.22\textwidth}
\centering
\includegraphics[width=\textwidth]{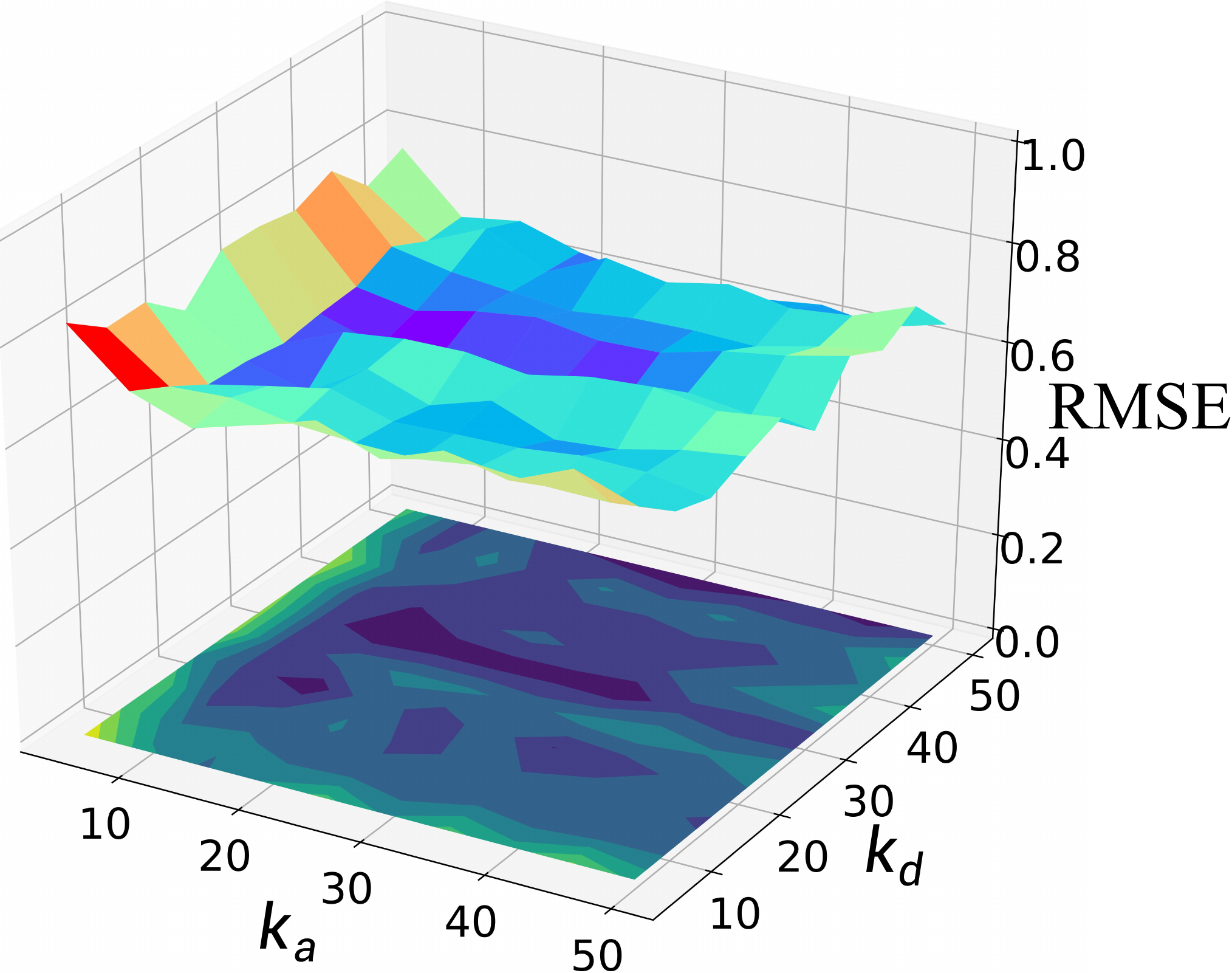}
\subcaption*{(b) RMSE on Epinions.}
\end{minipage}
\caption{Our DENC: parameter sensitivity of $k_{a}$ and $k_{d}$ against (a) MAE (b) RMSE on \texttt{Ciao} and \texttt{Epinions} dataset.}
\label{fig:parameter_sens}
\end{figure}

Figure~\ref{fig:parameter_sens} lays out the performance of DENC with different parameter settings.
For both datasets, the performance of our DENC is stable under different hyperparameters $k_a$ and $k_{d}$. 
The performance of DENC increases while the embedding size increase from approximately 0-15 for $k_{d}$; afterwards, its performance decreases.
It is clear that when the embedding size is set to approximately $k_a$=45 and $k_{d}$=15, our DENC method achieves the optimal performance.  
Our DENC is less sensitive to the change of $k_a$ than $k_{d}$, 
since MAE/RMSE values change with a obvious concave curve along $k_{d}$=0 to 50 in Figure~\ref{fig:parameter_sens}, while MAE/RMSE values only change gently with a downwards trend along $k_{a}$=0 to 50.
It is reasonable since $k_{d}$ controls the embedding size of disentangled user-item representation attained by the deconfounder model, i.e., the inherent factors,
while social network embedding size $k_{a}$ serves as the controller for auxiliary social information, 
the former can influence the essential user-item interaction while the latter affects the auxiliary information.

\subsection{Case Study (RQ4)}
We first investigate how the missing social relations affect the performance of DENC. 
We randomly mask a percentage of social relations to simulate the missing connections in social networks.
For \texttt{Epinions}, \texttt{Ciao} and \texttt{MovieLens} dataset, we fix the social network confounder as $\Delta(Z_u)=0$. Meanwhile, we exploit different percentages of missing social relations including \{20\%, 50\%, 80\%\}. Note that we do not consider the missing percentage of $100\%$, i.e., the social network information is completely unobserved.
Considering that the social network is viewed as a proxy variable of the confounder, the social network should provide partially known information. 
Following this guideline, we firstly investigate how the debias capability of our DENC method varies under the different missing percentages.
Secondly, we also report the ranking performance of DENC (percentages of missing social relations is set to $0\%$) under Precision@K and Recall@K with $K=\{10,15,20,25,30,35,40\}$ to evaluate our model thoroughly.

\begin{figure}[!htbp]
\centering
\begin{minipage}[t]{0.22\textwidth}
\centering
\includegraphics[width=\textwidth]{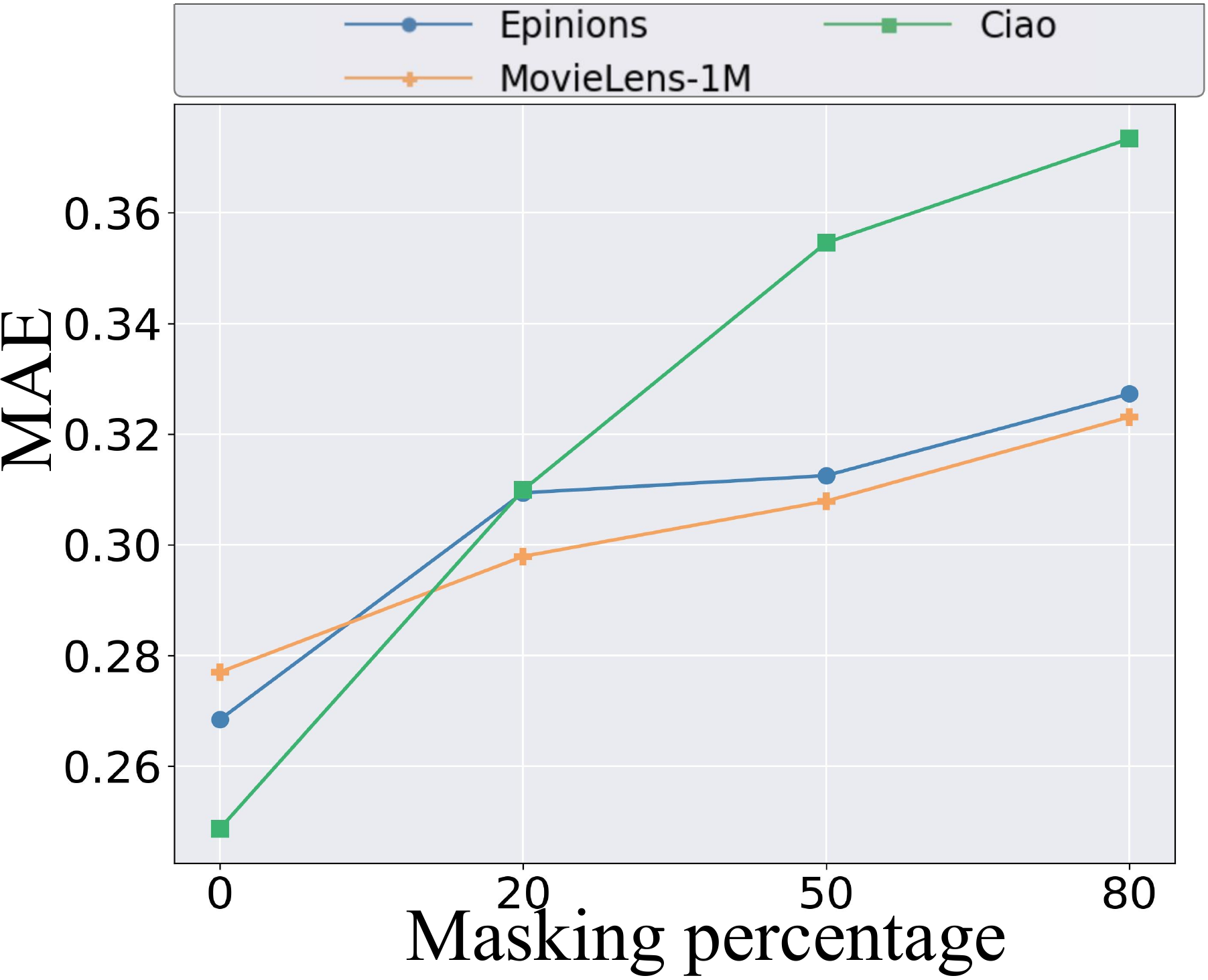}
\subcaption*{(a) MAE of DENC.}
\end{minipage}
\begin{minipage}[t]{0.22\textwidth}
\centering
\includegraphics[width=\textwidth]{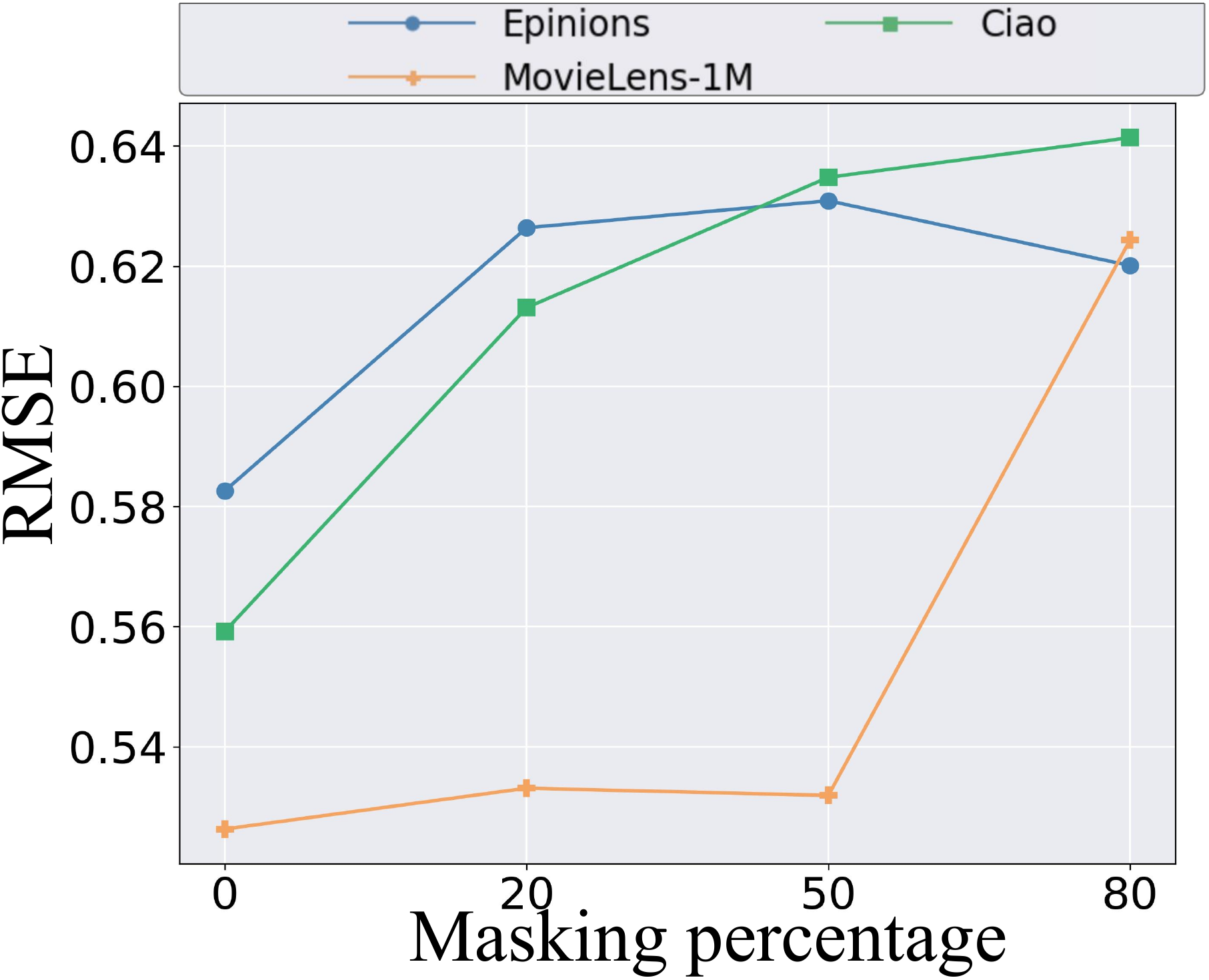}
\subcaption*{(b) RMSE of DENC.}
\end{minipage}
\begin{minipage}[t]{0.22\textwidth}
\centering
\includegraphics[width=\textwidth]{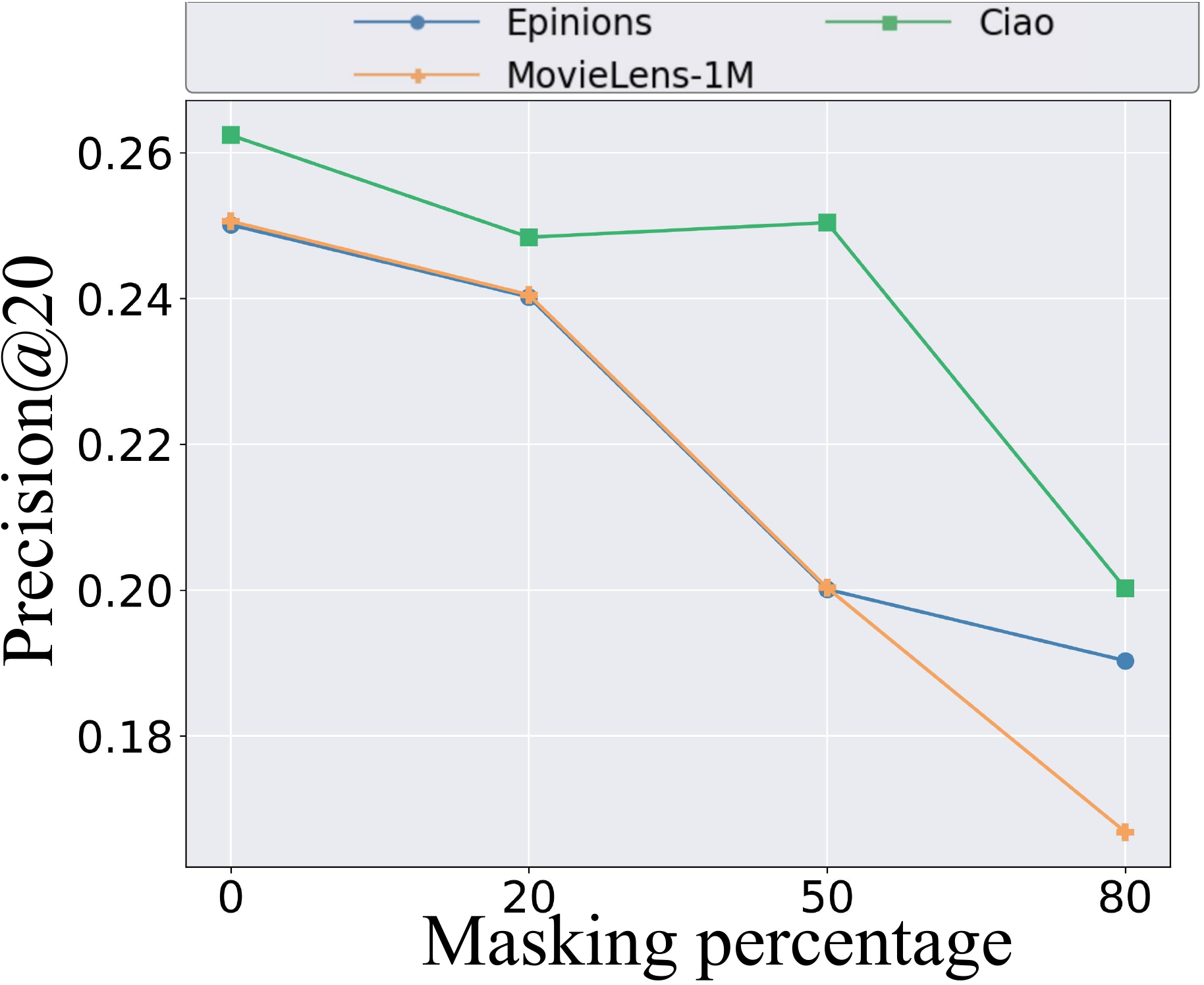}
\subcaption*{(c) Precision@20 of DENC.}
\end{minipage}
\begin{minipage}[t]{0.22\textwidth}
\centering
\includegraphics[width=\textwidth]{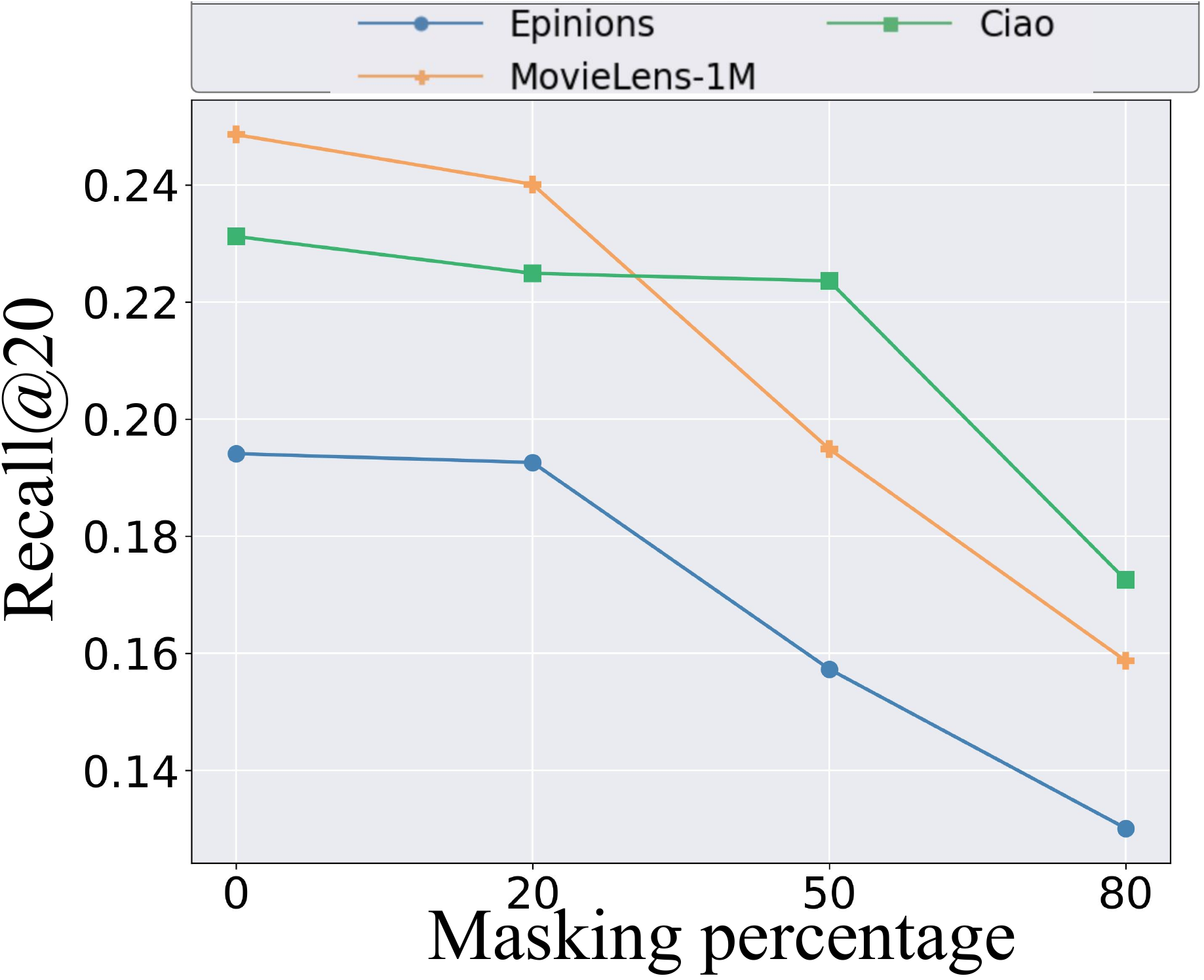}
\subcaption*{(d) Recall@20 of DENC.}
\end{minipage}
\caption{Our DENC: debias performance w.r.t. different missing percentages of social relation.}
\label{fig:masking}
\end{figure}

Figure~\ref{fig:masking} illustrates our debias performance w.r.t. different missing percentages of social relations on three datasets. As shown in Figure~\ref{fig:masking}, the missing social relations can obviously degrade the debias performance of DENC method. The performance evaluated by four metrics in Figure~\ref{fig:masking} consistently degrades when the missing percentage increases from 0\% to 80\%, which is consistent with the common observation. 
This indicates that the underlying social network can play a significant role in a recommendation, 
we consider the because it can capture the preference correlations between users and their neighbours.

\begin{figure}[htbp]
\centering
\begin{minipage}[t]{0.15\textwidth}
\centering
\includegraphics[width=\textwidth]{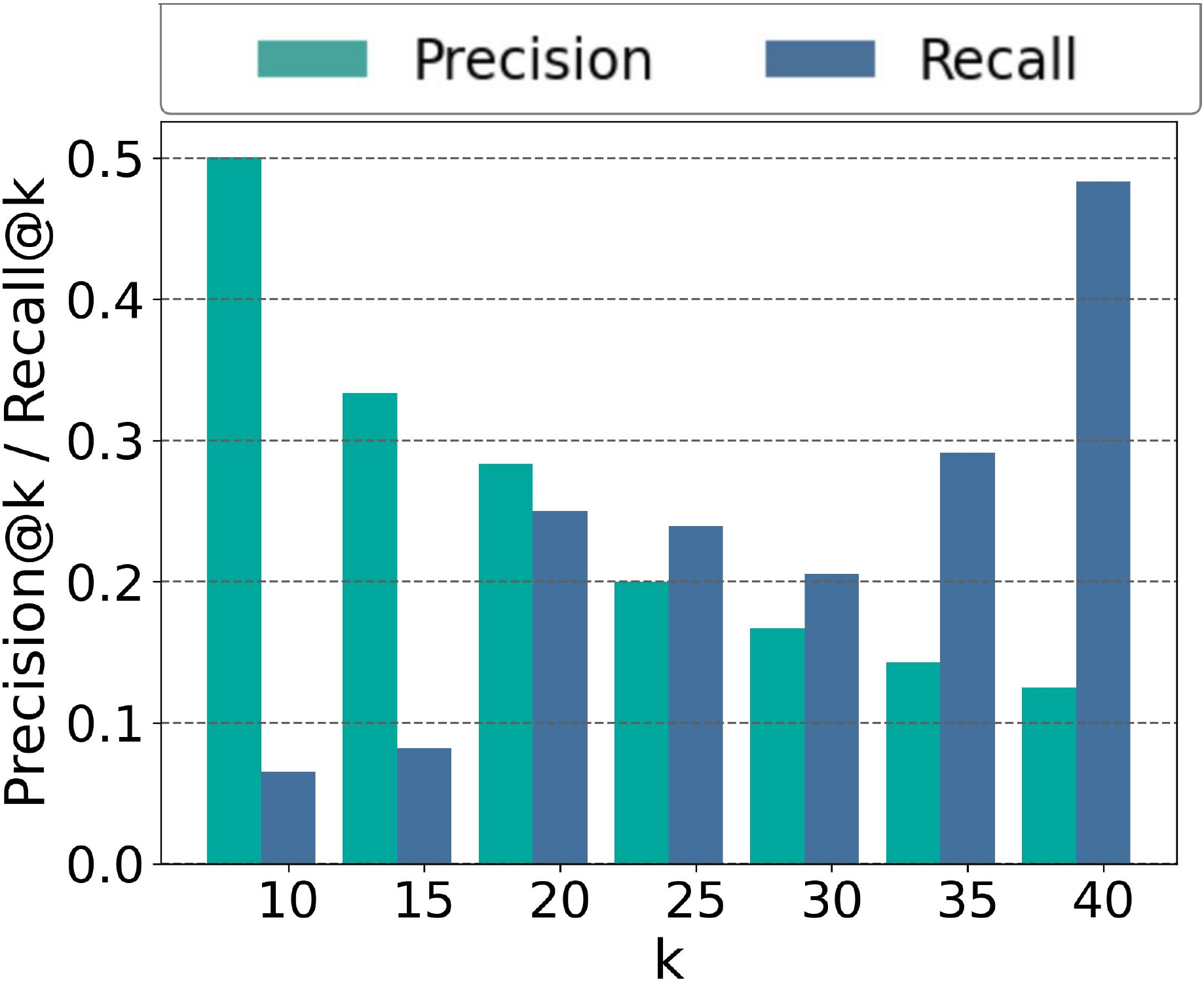}
\subcaption*{(a) \texttt{Epinions}.}
\end{minipage}
\begin{minipage}[t]{0.15\textwidth}
\centering
\includegraphics[width=\textwidth]{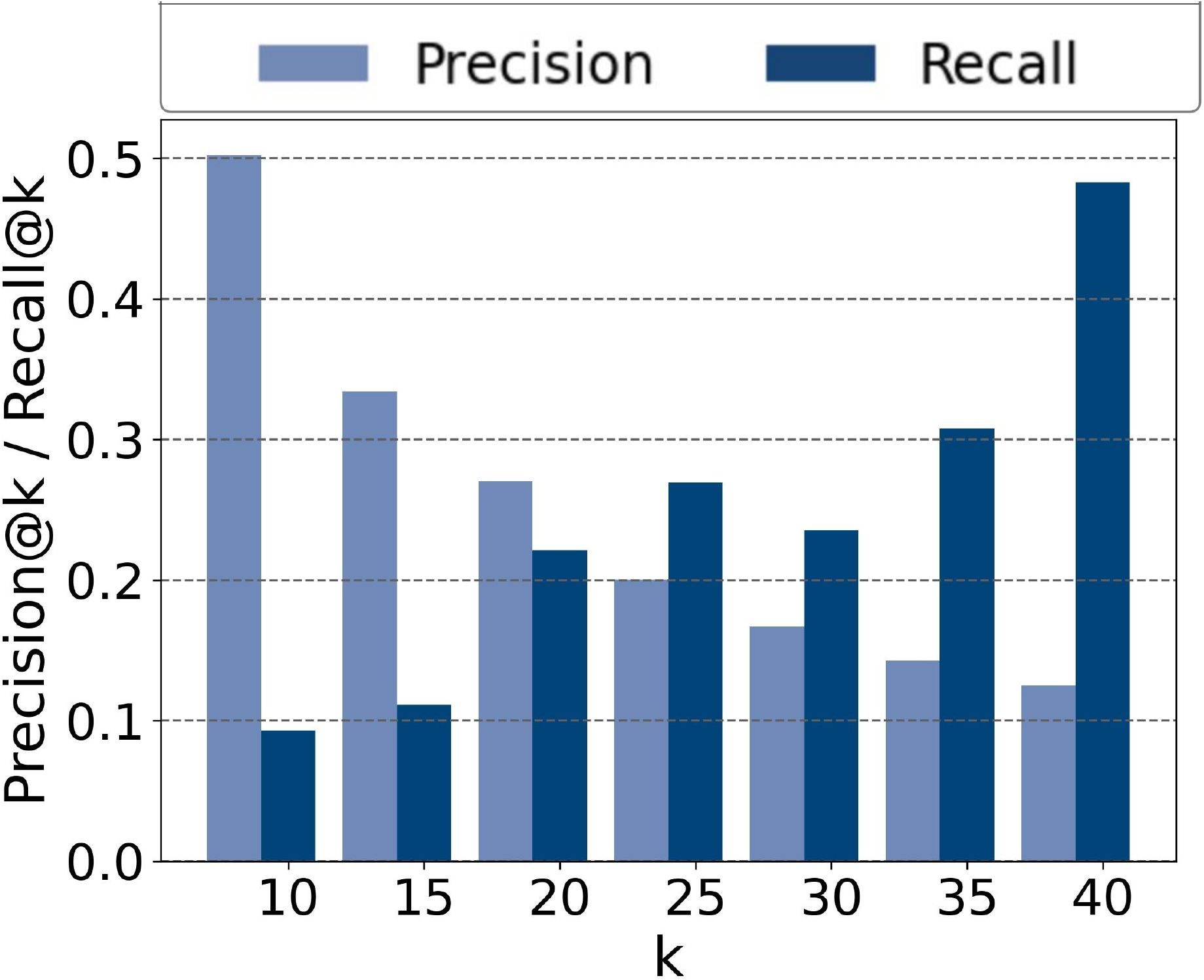}
\subcaption*{(b) \texttt{Ciao}.}
\end{minipage}
\begin{minipage}[t]{0.15\textwidth}
\centering
\includegraphics[width=\textwidth]{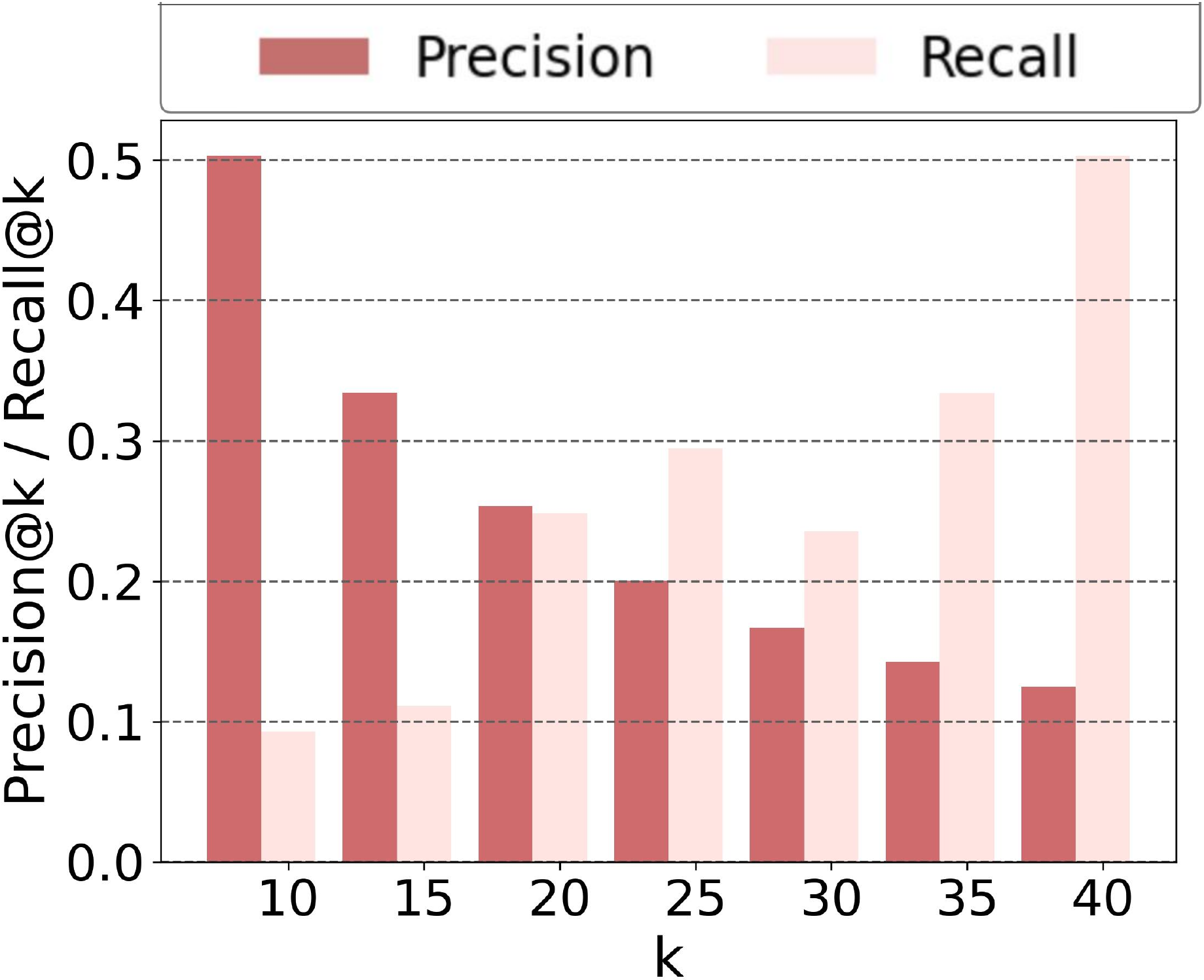}
\subcaption*{(c) \texttt{MovieLens-1M}\\($\Delta(Z_u)=0$).}
\end{minipage}
\caption{Performance of DENC in terms of Precision@K and RecallG@K under difference $K$}
\label{fig:pre_rec_k}
\end{figure}

Based on the evaluation on Precision@K and Recall@K, Figure~\ref{fig:pre_rec_k} show that DENC achieves stable performance on Top-$K$ recommendation when $K$ (i.e., the length of ranking list) varies from 10 to 40.
Our DENC can recommend more relevant items within top $K$ positions when the ranking list length increases.

\section{Conclusion and Future Work}
In this paper, we have researched the missing-not-at-random problem in the recommendation and addressed the confounding bias from a causal perspective.
Instead of merely relying on inherent information to account for selection bias, we developed a novel social network embedding based de-bias recommender for unbiased rating, through correcting the confounder effect arising from social networks.
We evaluate our DENC method on two real-world and one semi-synthetic recommendation datasets, 
with extensive experiments demonstrating the superiority of 
DENC in comparison to state-of-the-arts.
In future work, we will explore the effect of different exposure policies on the recommendation system using the intervention analysis in causal inference.
In addition, another promising further work is to explore the selection bias arisen from other confounder factors, e.g., user demographic features. This can be explained that a user's nationality affects which restaurant he is more likely to visit (i.e., exposure) and meanwhile affects how he will rate the restaurant (i.e., outcome).

\bibliographystyle{ACM-Reference-Format}
\bibliography{main}


\begin{thebibliography}{38}


\ifx \showCODEN    \undefined \def \showCODEN     #1{\unskip}     \fi
\ifx \showDOI      \undefined \def \showDOI       #1{#1}\fi
\ifx \showISBNx    \undefined \def \showISBNx     #1{\unskip}     \fi
\ifx \showISBNxiii \undefined \def \showISBNxiii  #1{\unskip}     \fi
\ifx \showISSN     \undefined \def \showISSN      #1{\unskip}     \fi
\ifx \showLCCN     \undefined \def \showLCCN      #1{\unskip}     \fi
\ifx \shownote     \undefined \def \shownote      #1{#1}          \fi
\ifx \showarticletitle \undefined \def \showarticletitle #1{#1}   \fi
\ifx \showURL      \undefined \def \showURL       {\relax}        \fi
\providecommand\bibfield[2]{#2}
\providecommand\bibinfo[2]{#2}
\providecommand\natexlab[1]{#1}
\providecommand\showeprint[2][]{arXiv:#2}

\bibitem[\protect\citeauthoryear{Bonner and Vasile}{Bonner and Vasile}{2018}]%
        {bonner2018causal}
\bibfield{author}{\bibinfo{person}{Stephen Bonner} {and}
  \bibinfo{person}{Flavian Vasile}.} \bibinfo{year}{2018}\natexlab{}.
\newblock \showarticletitle{Causal embeddings for recommendation}. In
  \bibinfo{booktitle}{\emph{Proceedings of the 12th ACM Conference on
  Recommender Systems}}. \bibinfo{pages}{104--112}.
\newblock


\bibitem[\protect\citeauthoryear{Bottou}{Bottou}{2010}]%
        {bottou2010large}
\bibfield{author}{\bibinfo{person}{L{\'e}on Bottou}.}
  \bibinfo{year}{2010}\natexlab{}.
\newblock \showarticletitle{Large-scale machine learning with stochastic
  gradient descent}.
\newblock In \bibinfo{booktitle}{\emph{Proceedings of COMPSTAT'2010}}.
  \bibinfo{publisher}{Springer}, \bibinfo{pages}{177--186}.
\newblock


\bibitem[\protect\citeauthoryear{Cui, Wang, Pei, and Zhu}{Cui
  et~al\mbox{.}}{2018}]%
        {cui2018survey}
\bibfield{author}{\bibinfo{person}{Peng Cui}, \bibinfo{person}{Xiao Wang},
  \bibinfo{person}{Jian Pei}, {and} \bibinfo{person}{Wenwu Zhu}.}
  \bibinfo{year}{2018}\natexlab{}.
\newblock \showarticletitle{A survey on network embedding}.
\newblock \bibinfo{journal}{\emph{IEEE Transactions on Knowledge and Data
  Engineering}} \bibinfo{volume}{31}, \bibinfo{number}{5}
  (\bibinfo{year}{2018}), \bibinfo{pages}{833--852}.
\newblock


\bibitem[\protect\citeauthoryear{Fan, Ma, Li, He, Zhao, Tang, and Yin}{Fan
  et~al\mbox{.}}{2019}]%
        {fan2019graph}
\bibfield{author}{\bibinfo{person}{Wenqi Fan}, \bibinfo{person}{Yao Ma},
  \bibinfo{person}{Qing Li}, \bibinfo{person}{Yuan He}, \bibinfo{person}{Eric
  Zhao}, \bibinfo{person}{Jiliang Tang}, {and} \bibinfo{person}{Dawei Yin}.}
  \bibinfo{year}{2019}\natexlab{}.
\newblock \showarticletitle{Graph neural networks for social recommendation}.
  In \bibinfo{booktitle}{\emph{The World Wide Web Conference}}.
  \bibinfo{pages}{417--426}.
\newblock


\bibitem[\protect\citeauthoryear{Glorot and Bengio}{Glorot and Bengio}{2010}]%
        {glorot2010understanding}
\bibfield{author}{\bibinfo{person}{Xavier Glorot} {and} \bibinfo{person}{Yoshua
  Bengio}.} \bibinfo{year}{2010}\natexlab{}.
\newblock \showarticletitle{Understanding the difficulty of training deep
  feedforward neural networks}. In \bibinfo{booktitle}{\emph{Proceedings of the
  thirteenth international conference on artificial intelligence and
  statistics}}. \bibinfo{pages}{249--256}.
\newblock


\bibitem[\protect\citeauthoryear{Grover and Leskovec}{Grover and
  Leskovec}{2016}]%
        {grover2016node2vec}
\bibfield{author}{\bibinfo{person}{Aditya Grover} {and} \bibinfo{person}{Jure
  Leskovec}.} \bibinfo{year}{2016}\natexlab{}.
\newblock \showarticletitle{node2vec: Scalable feature learning for networks}.
  In \bibinfo{booktitle}{\emph{Proceedings of the 22nd ACM SIGKDD international
  conference on Knowledge discovery and data mining}}. ACM,
  \bibinfo{pages}{855--864}.
\newblock


\bibitem[\protect\citeauthoryear{Guo, Tang, Ye, Li, and He}{Guo
  et~al\mbox{.}}{[n. d.]}]%
        {guo1703factorization}
\bibfield{author}{\bibinfo{person}{H Guo}, \bibinfo{person}{R Tang},
  \bibinfo{person}{Y Ye}, \bibinfo{person}{Z Li}, {and}
  \bibinfo{person}{X~DeepFM He}.} \bibinfo{year}{[n. d.]}\natexlab{}.
\newblock \showarticletitle{a factorization-machine based neural network for
  CTR prediction. arXiv 2017}.
\newblock \bibinfo{journal}{\emph{arXiv preprint arXiv:1703.04247}}
  (\bibinfo{year}{[n. d.]}).
\newblock


\bibitem[\protect\citeauthoryear{He, Zhang, Kan, and Chua}{He
  et~al\mbox{.}}{2016}]%
        {he2016fast}
\bibfield{author}{\bibinfo{person}{Xiangnan He}, \bibinfo{person}{Hanwang
  Zhang}, \bibinfo{person}{Min-Yen Kan}, {and} \bibinfo{person}{Tat-Seng
  Chua}.} \bibinfo{year}{2016}\natexlab{}.
\newblock \showarticletitle{Fast matrix factorization for online recommendation
  with implicit feedback}. In \bibinfo{booktitle}{\emph{Proceedings of the 39th
  International ACM SIGIR conference on Research and Development in Information
  Retrieval}}. \bibinfo{pages}{549--558}.
\newblock


\bibitem[\protect\citeauthoryear{Henderson, Gallagher, Eliassi-Rad, Tong, Basu,
  Akoglu, Koutra, Faloutsos, and Li}{Henderson et~al\mbox{.}}{2012}]%
        {henderson2012rolx}
\bibfield{author}{\bibinfo{person}{Keith Henderson}, \bibinfo{person}{Brian
  Gallagher}, \bibinfo{person}{Tina Eliassi-Rad}, \bibinfo{person}{Hanghang
  Tong}, \bibinfo{person}{Sugato Basu}, \bibinfo{person}{Leman Akoglu},
  \bibinfo{person}{Danai Koutra}, \bibinfo{person}{Christos Faloutsos}, {and}
  \bibinfo{person}{Lei Li}.} \bibinfo{year}{2012}\natexlab{}.
\newblock \showarticletitle{Rolx: structural role extraction \& mining in large
  graphs}. In \bibinfo{booktitle}{\emph{Proceedings of the 18th ACM SIGKDD
  international conference on Knowledge discovery and data mining}}.
  \bibinfo{pages}{1231--1239}.
\newblock


\bibitem[\protect\citeauthoryear{Hern{\'a}ndez-Lobato, Houlsby, and
  Ghahramani}{Hern{\'a}ndez-Lobato et~al\mbox{.}}{2014}]%
        {hernandez2014probabilistic}
\bibfield{author}{\bibinfo{person}{Jos{\'e}~Miguel Hern{\'a}ndez-Lobato},
  \bibinfo{person}{Neil Houlsby}, {and} \bibinfo{person}{Zoubin Ghahramani}.}
  \bibinfo{year}{2014}\natexlab{}.
\newblock \showarticletitle{Probabilistic matrix factorization with non-random
  missing data}. In \bibinfo{booktitle}{\emph{International Conference on
  Machine Learning}}. \bibinfo{pages}{1512--1520}.
\newblock


\bibitem[\protect\citeauthoryear{Hu, Koren, and Volinsky}{Hu
  et~al\mbox{.}}{2008}]%
        {hu2008collaborative}
\bibfield{author}{\bibinfo{person}{Yifan Hu}, \bibinfo{person}{Yehuda Koren},
  {and} \bibinfo{person}{Chris Volinsky}.} \bibinfo{year}{2008}\natexlab{}.
\newblock \showarticletitle{Collaborative filtering for implicit feedback
  datasets}. In \bibinfo{booktitle}{\emph{2008 Eighth IEEE International
  Conference on Data Mining}}. Ieee, \bibinfo{pages}{263--272}.
\newblock


\bibitem[\protect\citeauthoryear{Jamali and Ester}{Jamali and Ester}{2010}]%
        {jamali2010matrix}
\bibfield{author}{\bibinfo{person}{Mohsen Jamali} {and} \bibinfo{person}{Martin
  Ester}.} \bibinfo{year}{2010}\natexlab{}.
\newblock \showarticletitle{A matrix factorization technique with trust
  propagation for recommendation in social networks}. In
  \bibinfo{booktitle}{\emph{Proceedings of the fourth ACM conference on
  Recommender systems}}. \bibinfo{pages}{135--142}.
\newblock


\bibitem[\protect\citeauthoryear{Koren}{Koren}{2008}]%
        {koren2008factorization}
\bibfield{author}{\bibinfo{person}{Yehuda Koren}.}
  \bibinfo{year}{2008}\natexlab{}.
\newblock \showarticletitle{Factorization meets the neighborhood: a
  multifaceted collaborative filtering model}. In
  \bibinfo{booktitle}{\emph{Proceedings of the 14th ACM SIGKDD international
  conference on Knowledge discovery and data mining}}.
  \bibinfo{pages}{426--434}.
\newblock


\bibitem[\protect\citeauthoryear{Koren and Bell}{Koren and Bell}{[n. d.]}]%
        {koren2015advances}
\bibfield{author}{\bibinfo{person}{Yehuda Koren} {and} \bibinfo{person}{Robert
  Bell}.} \bibinfo{year}{[n. d.]}\natexlab{}.
\newblock \showarticletitle{Advances in collaborative filtering}.
\newblock In \bibinfo{booktitle}{\emph{Recommender systems handbook}}.
  \bibinfo{publisher}{Springer}, \bibinfo{pages}{77--118}.
\newblock


\bibitem[\protect\citeauthoryear{Li, Wang, Ren, Bing, and Lam}{Li
  et~al\mbox{.}}{2017b}]%
        {li2017neural}
\bibfield{author}{\bibinfo{person}{Piji Li}, \bibinfo{person}{Zihao Wang},
  \bibinfo{person}{Zhaochun Ren}, \bibinfo{person}{Lidong Bing}, {and}
  \bibinfo{person}{Wai Lam}.} \bibinfo{year}{2017}\natexlab{b}.
\newblock \showarticletitle{Neural rating regression with abstractive tips
  generation for recommendation}. In \bibinfo{booktitle}{\emph{Proceedings of
  the 40th International ACM SIGIR conference on Research and Development in
  Information Retrieval}}. \bibinfo{pages}{345--354}.
\newblock


\bibitem[\protect\citeauthoryear{Li, Gao, Rong, Wen, Xiong, Jia, and Dou}{Li
  et~al\mbox{.}}{2017a}]%
        {li2017social}
\bibfield{author}{\bibinfo{person}{Wentao Li}, \bibinfo{person}{Min Gao},
  \bibinfo{person}{Wenge Rong}, \bibinfo{person}{Junhao Wen},
  \bibinfo{person}{Qingyu Xiong}, \bibinfo{person}{Ruixi Jia}, {and}
  \bibinfo{person}{Tong Dou}.} \bibinfo{year}{2017}\natexlab{a}.
\newblock \showarticletitle{Social recommendation using Euclidean embedding}.
  In \bibinfo{booktitle}{\emph{2017 International Joint Conference on Neural
  Networks (IJCNN)}}. IEEE, \bibinfo{pages}{589--595}.
\newblock


\bibitem[\protect\citeauthoryear{Liang, Charlin, and Blei}{Liang
  et~al\mbox{.}}{2016}]%
        {liang2016causal}
\bibfield{author}{\bibinfo{person}{Dawen Liang}, \bibinfo{person}{Laurent
  Charlin}, {and} \bibinfo{person}{David~M Blei}.}
  \bibinfo{year}{2016}\natexlab{}.
\newblock \showarticletitle{Causal inference for recommendation}. In
  \bibinfo{booktitle}{\emph{Causation: Foundation to Application, Workshop at
  UAI}}. AUAI.
\newblock


\bibitem[\protect\citeauthoryear{Lim, McAuley, and Lanckriet}{Lim
  et~al\mbox{.}}{2015}]%
        {lim2015top}
\bibfield{author}{\bibinfo{person}{Daryl Lim}, \bibinfo{person}{Julian
  McAuley}, {and} \bibinfo{person}{Gert Lanckriet}.}
  \bibinfo{year}{2015}\natexlab{}.
\newblock \showarticletitle{Top-n recommendation with missing implicit
  feedback}. In \bibinfo{booktitle}{\emph{Proceedings of the 9th ACM Conference
  on Recommender Systems}}. \bibinfo{pages}{309--312}.
\newblock


\bibitem[\protect\citeauthoryear{Ma, Zhou, Liu, Lyu, and King}{Ma
  et~al\mbox{.}}{2011}]%
        {ma2011recommender}
\bibfield{author}{\bibinfo{person}{Hao Ma}, \bibinfo{person}{Dengyong Zhou},
  \bibinfo{person}{Chao Liu}, \bibinfo{person}{Michael~R Lyu}, {and}
  \bibinfo{person}{Irwin King}.} \bibinfo{year}{2011}\natexlab{}.
\newblock \showarticletitle{Recommender systems with social regularization}. In
  \bibinfo{booktitle}{\emph{Proceedings of the fourth ACM international
  conference on Web search and data mining}}. \bibinfo{pages}{287--296}.
\newblock


\bibitem[\protect\citeauthoryear{Marlin and Zemel}{Marlin and Zemel}{2009}]%
        {marlin2009collaborative}
\bibfield{author}{\bibinfo{person}{Benjamin~M Marlin} {and}
  \bibinfo{person}{Richard~S Zemel}.} \bibinfo{year}{2009}\natexlab{}.
\newblock \showarticletitle{Collaborative prediction and ranking with
  non-random missing data}. In \bibinfo{booktitle}{\emph{Proceedings of the
  third ACM conference on Recommender systems}}. \bibinfo{pages}{5--12}.
\newblock


\bibitem[\protect\citeauthoryear{Mnih and Salakhutdinov}{Mnih and
  Salakhutdinov}{2008}]%
        {mnih2008probabilistic}
\bibfield{author}{\bibinfo{person}{Andriy Mnih} {and} \bibinfo{person}{Russ~R
  Salakhutdinov}.} \bibinfo{year}{2008}\natexlab{}.
\newblock \showarticletitle{Probabilistic matrix factorization}. In
  \bibinfo{booktitle}{\emph{Advances in neural information processing
  systems}}. \bibinfo{pages}{1257--1264}.
\newblock


\bibitem[\protect\citeauthoryear{M{\"u}ller}{M{\"u}ller}{1997}]%
        {muller1997integral}
\bibfield{author}{\bibinfo{person}{Alfred M{\"u}ller}.}
  \bibinfo{year}{1997}\natexlab{}.
\newblock \showarticletitle{Integral probability metrics and their generating
  classes of functions}.
\newblock \bibinfo{journal}{\emph{Advances in Applied Probability}}
  (\bibinfo{year}{1997}), \bibinfo{pages}{429--443}.
\newblock


\bibitem[\protect\citeauthoryear{Pan, Zhou, Cao, Liu, Lukose, Scholz, and
  Yang}{Pan et~al\mbox{.}}{2008}]%
        {pan2008one}
\bibfield{author}{\bibinfo{person}{Rong Pan}, \bibinfo{person}{Yunhong Zhou},
  \bibinfo{person}{Bin Cao}, \bibinfo{person}{Nathan~N Liu},
  \bibinfo{person}{Rajan Lukose}, \bibinfo{person}{Martin Scholz}, {and}
  \bibinfo{person}{Qiang Yang}.} \bibinfo{year}{2008}\natexlab{}.
\newblock \showarticletitle{One-class collaborative filtering}. In
  \bibinfo{booktitle}{\emph{2008 Eighth IEEE International Conference on Data
  Mining}}. IEEE, \bibinfo{pages}{502--511}.
\newblock


\bibitem[\protect\citeauthoryear{Pearl}{Pearl}{2009}]%
        {pearl2009causality}
\bibfield{author}{\bibinfo{person}{Judea Pearl}.}
  \bibinfo{year}{2009}\natexlab{}.
\newblock \bibinfo{booktitle}{\emph{Causality}}.
\newblock \bibinfo{publisher}{Cambridge university press}.
\newblock


\bibitem[\protect\citeauthoryear{Purushotham, Liu, and Kuo}{Purushotham
  et~al\mbox{.}}{2012}]%
        {purushotham2012collaborative}
\bibfield{author}{\bibinfo{person}{Sanjay Purushotham}, \bibinfo{person}{Yan
  Liu}, {and} \bibinfo{person}{C-C~Jay Kuo}.} \bibinfo{year}{2012}\natexlab{}.
\newblock \showarticletitle{Collaborative topic regression with social matrix
  factorization for recommendation systems}.
\newblock \bibinfo{journal}{\emph{arXiv preprint arXiv:1206.4684}}
  (\bibinfo{year}{2012}).
\newblock


\bibitem[\protect\citeauthoryear{Rendle and Schmidt-Thieme}{Rendle and
  Schmidt-Thieme}{2008}]%
        {rendle2008online}
\bibfield{author}{\bibinfo{person}{Steffen Rendle} {and} \bibinfo{person}{Lars
  Schmidt-Thieme}.} \bibinfo{year}{2008}\natexlab{}.
\newblock \showarticletitle{Online-updating regularized kernel matrix
  factorization models for large-scale recommender systems}. In
  \bibinfo{booktitle}{\emph{Proceedings of the 2008 ACM conference on
  Recommender systems}}. \bibinfo{pages}{251--258}.
\newblock


\bibitem[\protect\citeauthoryear{Rubin}{Rubin}{1974}]%
        {rubin1974estimating}
\bibfield{author}{\bibinfo{person}{Donald~B Rubin}.}
  \bibinfo{year}{1974}\natexlab{}.
\newblock \showarticletitle{Estimating causal effects of treatments in
  randomized and nonrandomized studies.}
\newblock \bibinfo{journal}{\emph{Journal of educational Psychology}}
  \bibinfo{volume}{66}, \bibinfo{number}{5} (\bibinfo{year}{1974}),
  \bibinfo{pages}{688}.
\newblock


\bibitem[\protect\citeauthoryear{Saito}{Saito}{2020}]%
        {saito2020asymmetric}
\bibfield{author}{\bibinfo{person}{Yuta Saito}.}
  \bibinfo{year}{2020}\natexlab{}.
\newblock \showarticletitle{Asymmetric Tri-training for Debiasing
  Missing-Not-At-Random Explicit Feedback}. In
  \bibinfo{booktitle}{\emph{Proceedings of the 43rd International ACM SIGIR
  Conference on Research and Development in Information Retrieval}}.
  \bibinfo{pages}{309--318}.
\newblock


\bibitem[\protect\citeauthoryear{Schnabel, Swaminathan, Singh, Chandak, and
  Joachims}{Schnabel et~al\mbox{.}}{2016}]%
        {schnabel2016recommendations}
\bibfield{author}{\bibinfo{person}{Tobias Schnabel}, \bibinfo{person}{Adith
  Swaminathan}, \bibinfo{person}{Ashudeep Singh}, \bibinfo{person}{Navin
  Chandak}, {and} \bibinfo{person}{Thorsten Joachims}.}
  \bibinfo{year}{2016}\natexlab{}.
\newblock \showarticletitle{Recommendations as treatments: Debiasing learning
  and evaluation}.
\newblock \bibinfo{journal}{\emph{arXiv preprint arXiv:1602.05352}}
  (\bibinfo{year}{2016}).
\newblock


\bibitem[\protect\citeauthoryear{Shalit, Johansson, and Sontag}{Shalit
  et~al\mbox{.}}{2017}]%
        {shalit2017estimating}
\bibfield{author}{\bibinfo{person}{Uri Shalit}, \bibinfo{person}{Fredrik~D
  Johansson}, {and} \bibinfo{person}{David Sontag}.}
  \bibinfo{year}{2017}\natexlab{}.
\newblock \showarticletitle{Estimating individual treatment effect:
  generalization bounds and algorithms}. In
  \bibinfo{booktitle}{\emph{International Conference on Machine Learning}}.
  PMLR, \bibinfo{pages}{3076--3085}.
\newblock


\bibitem[\protect\citeauthoryear{Sportisse, Boyer, and Josse}{Sportisse
  et~al\mbox{.}}{2020}]%
        {sportisse2020imputation}
\bibfield{author}{\bibinfo{person}{Aude Sportisse}, \bibinfo{person}{Claire
  Boyer}, {and} \bibinfo{person}{Julie Josse}.}
  \bibinfo{year}{2020}\natexlab{}.
\newblock \showarticletitle{Imputation and low-rank estimation with Missing Not
  At Random data}.
\newblock \bibinfo{journal}{\emph{Statistics and Computing}}
  \bibinfo{volume}{30}, \bibinfo{number}{6} (\bibinfo{year}{2020}),
  \bibinfo{pages}{1629--1643}.
\newblock


\bibitem[\protect\citeauthoryear{Tang, Gao, and Liu}{Tang
  et~al\mbox{.}}{2012}]%
        {tang-etal12a}
\bibfield{author}{\bibinfo{person}{J. Tang}, \bibinfo{person}{H. Gao}, {and}
  \bibinfo{person}{H. Liu}.} \bibinfo{year}{2012}\natexlab{}.
\newblock \showarticletitle{m{T}rust: {D}iscerning multi-faceted trust in a
  connected world}. In \bibinfo{booktitle}{\emph{Proceedings of the fifth ACM
  international conference on Web search and data mining}}. ACM,
  \bibinfo{pages}{93--102}.
\newblock


\bibitem[\protect\citeauthoryear{Tang, Qu, Wang, Zhang, Yan, and Mei}{Tang
  et~al\mbox{.}}{2015}]%
        {tang2015line}
\bibfield{author}{\bibinfo{person}{Jian Tang}, \bibinfo{person}{Meng Qu},
  \bibinfo{person}{Mingzhe Wang}, \bibinfo{person}{Ming Zhang},
  \bibinfo{person}{Jun Yan}, {and} \bibinfo{person}{Qiaozhu Mei}.}
  \bibinfo{year}{2015}\natexlab{}.
\newblock \showarticletitle{Line: Large-scale information network embedding}.
  In \bibinfo{booktitle}{\emph{Proceedings of the 24th international conference
  on world wide web}}. \bibinfo{pages}{1067--1077}.
\newblock


\bibitem[\protect\citeauthoryear{Wang, Cui, and Zhu}{Wang
  et~al\mbox{.}}{2016}]%
        {wang2016structural}
\bibfield{author}{\bibinfo{person}{Daixin Wang}, \bibinfo{person}{Peng Cui},
  {and} \bibinfo{person}{Wenwu Zhu}.} \bibinfo{year}{2016}\natexlab{}.
\newblock \showarticletitle{Structural deep network embedding}. In
  \bibinfo{booktitle}{\emph{Proceedings of the 22nd ACM SIGKDD international
  conference on Knowledge discovery and data mining}}.
  \bibinfo{pages}{1225--1234}.
\newblock


\bibitem[\protect\citeauthoryear{Wang, Zhang, Sun, and Qi}{Wang
  et~al\mbox{.}}{2019}]%
        {wang2019doubly}
\bibfield{author}{\bibinfo{person}{Xiaojie Wang}, \bibinfo{person}{Rui Zhang},
  \bibinfo{person}{Yu Sun}, {and} \bibinfo{person}{Jianzhong Qi}.}
  \bibinfo{year}{2019}\natexlab{}.
\newblock \showarticletitle{Doubly robust joint learning for recommendation on
  data missing not at random}. In \bibinfo{booktitle}{\emph{International
  Conference on Machine Learning}}. \bibinfo{pages}{6638--6647}.
\newblock


\bibitem[\protect\citeauthoryear{Wang, Liang, Charlin, and Blei}{Wang
  et~al\mbox{.}}{2018}]%
        {wang2018deconfounded}
\bibfield{author}{\bibinfo{person}{Yixin Wang}, \bibinfo{person}{Dawen Liang},
  \bibinfo{person}{Laurent Charlin}, {and} \bibinfo{person}{David~M Blei}.}
  \bibinfo{year}{2018}\natexlab{}.
\newblock \showarticletitle{The deconfounded recommender: A causal inference
  approach to recommendation}.
\newblock \bibinfo{journal}{\emph{arXiv preprint arXiv:1808.06581}}
  (\bibinfo{year}{2018}).
\newblock


\bibitem[\protect\citeauthoryear{Yang, Lei, Liu, and Li}{Yang
  et~al\mbox{.}}{2016}]%
        {yang2016social}
\bibfield{author}{\bibinfo{person}{Bo Yang}, \bibinfo{person}{Yu Lei},
  \bibinfo{person}{Jiming Liu}, {and} \bibinfo{person}{Wenjie Li}.}
  \bibinfo{year}{2016}\natexlab{}.
\newblock \showarticletitle{Social collaborative filtering by trust}.
\newblock \bibinfo{journal}{\emph{IEEE transactions on pattern analysis and
  machine intelligence}} \bibinfo{volume}{39}, \bibinfo{number}{8}
  (\bibinfo{year}{2016}), \bibinfo{pages}{1633--1647}.
\newblock


\bibitem[\protect\citeauthoryear{Yang, Cui, Xuan, Wang, Belongie, and
  Estrin}{Yang et~al\mbox{.}}{2018}]%
        {yang2018unbiased}
\bibfield{author}{\bibinfo{person}{Longqi Yang}, \bibinfo{person}{Yin Cui},
  \bibinfo{person}{Yuan Xuan}, \bibinfo{person}{Chenyang Wang},
  \bibinfo{person}{Serge Belongie}, {and} \bibinfo{person}{Deborah Estrin}.}
  \bibinfo{year}{2018}\natexlab{}.
\newblock \showarticletitle{Unbiased offline recommender evaluation for
  missing-not-at-random implicit feedback}. In
  \bibinfo{booktitle}{\emph{Proceedings of the 12th ACM Conference on
  Recommender Systems}}. \bibinfo{pages}{279--287}.
\newblock


\end{thebibliography}

\newpage
\appendix

\section{APPENDIX}

\subsection{Datasets}\label{sec:data}
The statistics of baseline datasets are given in Table~\ref{tab:2}.
In \texttt{Epinions} and \texttt{Ciao}, the rating values are integers from 1 (like least) to 5 (like most). Since observed ratings are very sparse (rating density 0.0140\% for \texttt{Epinions} and 0.0368\% for \texttt{Ciao}), thus the rating prediction on these two datasets is challenging.

        In addition, we also simulate a semi-synthetic dataset based on \texttt{MovieLens}. It is well-known that \texttt{MovieLens} is a benchmark dataset of user-movie ratings without social network information.
For \texttt{MovieLens-1M}, we first need to construct a social network $G$ by placing an edge between each pair of users independently with a probability 0.5 depending on whether the nodes belong to $G$.
Recall that the social network is viewed as the confounder (common cause) which affects both exposure variables and ratings.
We generate the exposure assignment by the confounder $Z_u$ of three levels $\Delta(Z_u)\in\{-0.35,0,0.35\}$. Then, the exposure $a_{ui}$ and rating outcome $y_{ui}$ are simulated as follows.
\begin{flalign*}
a_{ui}&\sim\operatorname{Bern}\left(\Delta\left(Z_{u}\right)\right)\\
    y_{ui}&=a_{ui}\cdot(y_{ui}^{\text{mov}}+\beta_u\Delta\left(Z_{u}\right)+\varepsilon) &\varepsilon\sim N(0,1), \quad u\in G\\
    y_{ui}&=y_{ui}^{\text{mov}} &  u\notin G
    \label{eq:semi}
\end{flalign*}
where $y_{ui}^{\text{mov}}$ is the original rating in \texttt{MovieLens} and the parameter $\beta_u$ controls the amount of social network confounder.
The exposure $a_{ui}$ indicating whether item $i$ being exposed to user $u$ is given by a Bernoulli distribution parameterized by the confounder $Z_u$. 
The non-zero $a_{ui}$ is used to simulate the semi-synthetic rating $y_{ui}$ by the second equation. The third equation indicates that the ratings of user will keep unchanged if s/he is not connected by $G$.

\subsection{Baselines}\label{sec:baseline}
We compare our DENC against three groups of methods, covering matrix factorization method, social network-based method, and propensity-based method. For each group, we select its representative baselines with details as follows.

\begin{itemize}[leftmargin=*]
            \item \textbf{PMF}~\cite{mnih2008probabilistic}:
            The method utilizes user-item rating matrix and models latent factors of users and items by Gaussian distributions;
           
            \item \textbf{NRT}~\cite{li2017neural}: 
            A deep-learning method 
            that adopts multi-layer perceptron network to model user-item interactions for rating predictions.
            \item \textbf{SocialMF}~\cite{jamali2010matrix}:
            It considers the social information by adding the propagation of social relation into the matrix factorization model.
            \item \textbf{SoReg}~\cite{ma2011recommender}:
            It models social information as regularization terms to constrain the Matrix Factorization framework.
            \item \textbf{SREE}~\cite{li2017social}: 
            It models users and items embeddings into a Euclidean space as well as users' social relations.
            \item \textbf{GraphRec}~\cite{fan2019graph}:
            This is a state-of-the-art social recommender that models social information with Graph Neural Network, it 
            organizes user behaviors as a user-item interaction graph.
            \item \textbf{DeepFM}~\cite{guo1703factorization}\textbf{+}: 
            DeepFM is a state-of-the-art recommender that integrates Deep Neural Networks and Factorization Machine (FM).
            To incorporate the social information into DeepFM, 
            we change the output of FM in DeepFM+ to the linear combination of the original FM function in ~\cite{guo1703factorization} and
            the pre-trained \textit{node2vec} user embeddings.
            We also change the task of DeepMF from click-through rate (CTR) to rating prediction.
            \item \textbf{CausE}~\cite{bonner2018causal}:
            It firstly fits exposure variable embedding with Poisson factorization, then integrates the embedding into PMF for rating prediction.
            \item \textbf{D-WMF}~\cite{wang2018deconfounded}:
            A propensity-based model which uses Poisson Factorization to infer latent confounders then augments Weighted Matrix Factorization to correct for potential confounding bias.
        \end{itemize}
      
\subsection{Model Variants Configuration}
To get a better understanding of our DENC method,
we further evaluate the key components of DENC including \textit{Exposure model} and \textit{Social network confounder}. 
We evaluate the performance of DENC on the condition that if a specific component is removed, and then compare the performance of the intact DENC method.
In the following,  we define two variants of DENC as (1) DENC-$\alpha$ that removes \textit{Exposure model}; (2) DENC-$\beta$ that removes \textit{Social network confounder}.
Note that we do not consider the evaluation of removing \textit{Deconfounder} in DENC, since \textit{Deconfounder} models the inherent factors of user-item information, removing user-item information in a recommender can result in poor performance.
We record evaluation results in Table~\ref{tab:component} and have the following findings:
\begin{itemize}[leftmargin=*]
\item By comparing DENC with DENC-$\alpha$, we find that \textit{Exposure model} is important for capturing missing patterns and thus boosting the recommendation performance. Removing \textit{Exposure model} can lead a drastic degradation of MAE/RMSE by 20.41\%/24.08\% on \texttt{Epinions} and 18.93\%/24.34\% on \texttt{Ciao}, respectively. 
\item We observe that without \textit{Social network confounder}, the performance of DENC-$\beta$ is deteriorated significantly, with the degradation of MAE/RMSE by 16.10\%/20.50\% on \texttt{Epinions} and 13.83\%/11.31\% on \texttt{Ciao}, respectively. 
\item \textit{Exposure model} has a greater impact on DENC compared with \textit{Social network confounder}. 
It is reasonable since \textit{Exposure model} simulates the missing patterns, then \textit{Social network confounder} can consequently debias the potential confounding bias under the guidance of missing patterns.
\end{itemize}

\begin{table}
\centering
  \caption{
  Experimental results of DENC-$\alpha$ and DENC-$\beta$.
  }

\begin{tabular}{c||c||cc}\hline
    Dataset &Models &MAE &RMSE\\\hhline{-||-||--}
    \multicolumn{1}{c||}{\texttt{Epinions}} &DENC-$\alpha$ &0.4725 &0.8234 \\
    \multicolumn{1}{c||}{} &DENC-$\beta$ &0.4294 &0.7876 \\
    \multicolumn{1}{c||}{} &DENC &0.2684 &0.5826\\
    \multicolumn{1}{c||}{\texttt{Ciao}} &DENC-$\alpha$ &0.4380 &0.8026 \\
    \multicolumn{1}{c||}{} &DENC-$\beta$ &0.3870 &0.6723 \\
    \multicolumn{1}{c||}{} &DENC &0.2487 &0.5592\\
    \hline
\end{tabular}
\label{tab:component}
\end{table}

\subsection{Investigation on Different Network Embedding Methods}\label{sec:embedding}
We construct network embedding with  node2vec~\cite{grover2016node2vec} that has the capacity of learning richer representations by adding flexibility in exploring neighborhoods of nodes. 
Besides, by adjusting the weight of the random walk between breadth-first and depth-first sampling, embeddings generated by node2vec can balance the trade-off between homophily and structural equivalence~\cite{henderson2012rolx}, both of which are essential feature expressions in recommendation systems.
The key characteristic of node2vec is its scalability and efficiency as it scales to networks of millions of nodes.

By comparison, we further investigate how different network embedding methods impact the performance of DENC, i.e., LINE~\cite{tang2015line}, SDNE~\cite{wang2016structural}.
\begin{itemize}[leftmargin=*]
\item \textbf{LINE}~\cite{tang2015line} preserves both first-order and second-order proximities, it suits arbitrary types of information networks and can easily scale to millions of nodes.
\item \textbf{SDNE}~\cite{wang2016structural} is a Deep Learning-based network embedding method, like LINE, it exploits the first-order and second-order proximity jointly to preserve the network structure.
\end{itemize}
We train the three embedding methods with embedding size $d$=10 while the batch size and epochs are set to 1024 and 50, respectively. The experimental results are given in
Table~\ref{tab:emb}. 

\begin{table}[!htb]
\centering
  \caption{
  Experimental results of DENC under node2vec, LINE, SDNE.
  }
\resizebox{0.48\textwidth}{!}{
\begin{tabular}{c||c||cc||cc}\hline
    Dataset &Embedding &MAE &RMSE &Precision@20 &Recall@20\\\hhline{-||-||--||--}
    \multicolumn{1}{c||}{\texttt{Epinions}} &node2vec &0.2684 &0.5826 &0.2832 &0.2501\\
    \multicolumn{1}{c||}{} &LINE &0.4241 &0.6307 &0.1736 &0.1534\\
    \multicolumn{1}{c||}{} &SDNE &0.4021 &0.6137 &0.1928 &0.1837\\
    \multicolumn{1}{c||}{\texttt{Ciao}} &node2vec &0.2487 &0.5592 & 0.2703 & 0.2212\\
    \multicolumn{1}{c||}{} &LINE &0.5218 &0.7605 &0.1504 &0.1209 \\
    \multicolumn{1}{c||}{} &SDNE &0.4538 &0.6274 &0.2082 &0.1594\\
    \hline
\end{tabular}
}
\label{tab:emb}
\end{table}

The results show that under the same experimental settings, DENC performs worse with embeddings trained by LINE and SDNE compared with node2vec on both datasets.
Although LINE considers the higher-order proximity, unlike node2vec, it still cannot balance the representation between homophily and structural equivalence~\cite{henderson2012rolx}, in which connectivity information and network structure information can be captured jointly. 
The results show that our DENC benefits more from the balanced representation that can learn both the connectivity information and network structure information.
Based on higher-order proximity, SDNE develops a deep-learning representation method. However, compared with node2vec, SDNE suffers from higher time complexity. 
The deep architecture of SDNE framework mainly causes the high time complexity of SDNE, the input vector dimension can expand to millions for the auto-encoder in SDNE~\cite{cui2018survey}.
Thus, we consider it reasonable that our DENC with SDNE embedding cannot outperform the counterpart with node2vec embedding under the same training epochs, since it requires more iterations for SDNE to get finer representation.

\end{document}